\documentclass[10pt,twocolumn,letterpaper]{article}

\usepackage[pagenumbers]{wacv} 

\definecolor{wacvblue}{rgb}{0.21,0.49,0.74}
\usepackage[pagebackref,breaklinks,colorlinks,allcolors=wacvblue]{hyperref}
\definecolor{blue}{rgb}{0,0,0} 

\def\wacvPaperID{758} 
\def\confName{WACV}
\def\confYear{2027}

\title{PRIMA: Boosting Animal Mesh Recovery with \\\textcolor{blue}{Taxonomic Priors} and Adaptation-Driven Data Bootstrapping
}

\author{
Xiaohang Yu,\quad Ti Wang,\quad Mackenzie Weygandt Mathis\\
\'Ecole Polytechnique F\'ed\'erale de Lausanne (EPFL)\\
{\tt\small \ mackenzie.mathis@epfl.ch}
}

\begin{document}
\maketitle
\begin{abstract}

Recovering articulated 3D animal meshes from monocular images is challenging due to the long-tailed distribution of species and limited large-scale 3D supervision. Existing methods rely on small annotated datasets and thus regress to average shapes and poses, leading to poor generalization to rare species and articulations.
We present PRIMA (\textbf{PRI}ors for \textbf{M}esh \textbf{A}daptation), a framework that tackles these limitations with two components: \textcolor{blue}{taxonomic shape priors} and adaptation-driven data bootstrapping. \textcolor{blue}{First, we derive taxonomic shape priors from BioCLIP image features, which are learned under taxonomic supervision, to inject species-level knowledge into mesh reconstruction, improving shape estimation across diverse quadrupeds.} Second, we propose an adaptation-driven bootstrapping scheme that turns abundant 2D animal annotations into reliable pseudo-3D supervision via lightweight instance-level mesh refinement and quality-aware filtering. This yields Quadruped3D, a large-scale pseudo-3D dataset constructed from real-world animal images, containing approximately 9K pseudo-3D annotations.
Together, these components alleviate the long-tail problem: extensive experiments on Animal3D, CtrlAni3D, Quadruped2D, and Animal Kingdom demonstrate state-of-the-art performance for mesh reconstruction, with particularly strong gains on challenging poses and underrepresented species. The code is available at \href{https://github.com/AdaptiveMotorControlLab/PRIMA}{https://github.com/AdaptiveMotorControlLab/PRIMA}. 

\end{abstract}

\section{Introduction}
\label{sec:intro}

Reconstructing articulated 3D animal meshes from a single image is a fundamental problem with applications in ecology, neuroscience, and biodiversity monitoring. Despite recent advances in monocular 3D reconstruction \cite{you2023co,dwivedi2024tokenhmr,pavlakos2024reconstructing,wu2024reconstructing}, animal mesh recovery remains particularly challenging due to extreme morphological diversity across species, highly variable articulation patterns, and the scarcity of large-scale 3D supervision \cite{li2025advances}.
\begin{figure}
\centering
\includegraphics[width=1.0\linewidth]{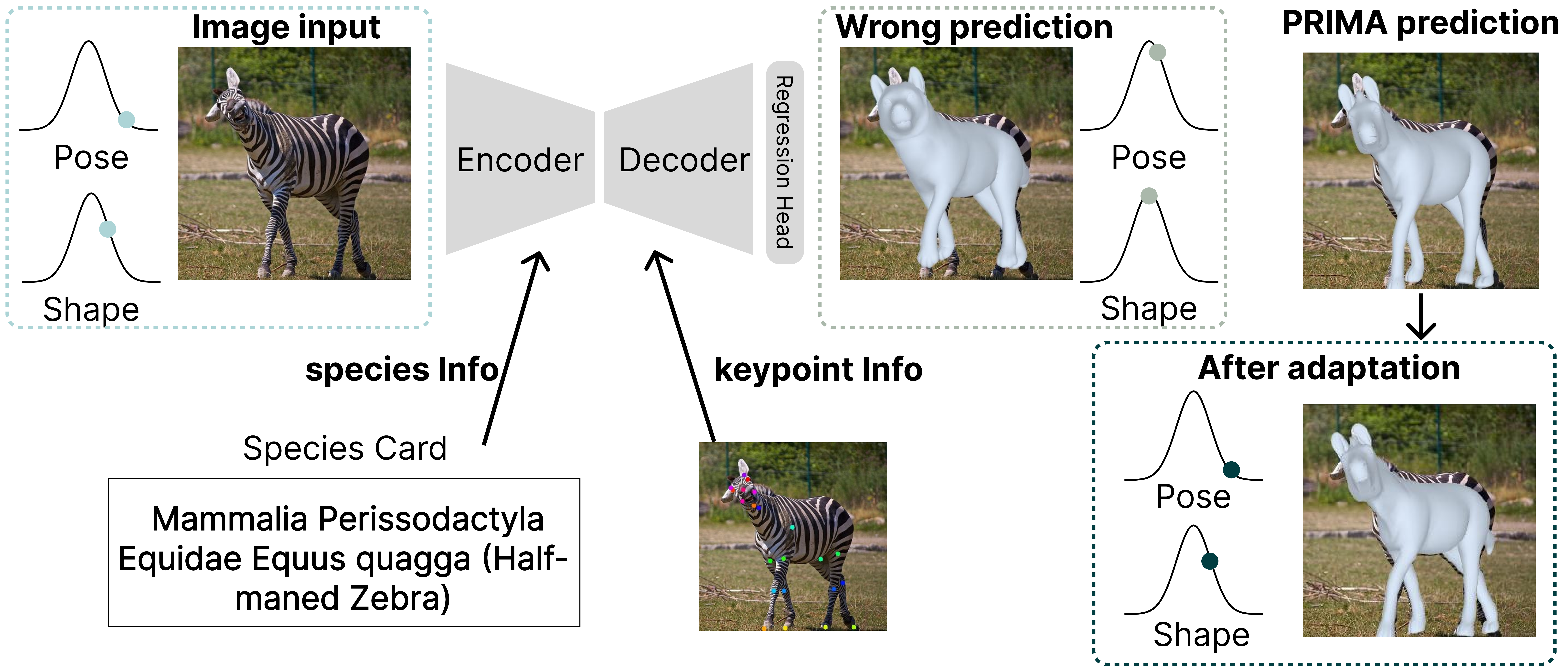}
\caption{\textbf{Overview of PRIMA.}
Existing animal mesh recovery methods suffer from long-tailed species distributions and limited 3D supervision, often producing reconstructions biased toward dominant families.
PRIMA addresses these challenges through two complementary mechanisms:
(1) BioCLIP-derived \textcolor{blue}{taxonomic shape priors} that provide species-aware shape conditioning, and
(2) adaptation-driven data bootstrapping that converts abundant 2D annotations into reliable pseudo-3D supervision.
Together, these components improve reconstruction fidelity and generalization across diverse quadruped species and poses.}
\label{fig:teaser}
\end{figure}
A central challenge underlying modern animal reconstruction systems is the long-tailed nature of available training data. Existing datasets provide reliable 3D supervision for only a limited set of species, while real-world imagery exhibits substantially broader species and pose diversity. Consequently, reconstruction models often learn biased priors dominated by overrepresented categories such as dogs and horses, resulting in poor generalization to rare species and unusual articulations.

To constrain this highly ambiguous problem, strong shape priors have proven essential \cite{loper2015smpl,pavlakos2019expressive,ferguson2025mhr}. Following the success of SMPL \cite{loper2015smpl} for humans, SMAL \cite{smalZuffi:CVPR:2017} introduced a unified articulated representation for quadruped animals and has become the foundation of modern animal mesh recovery methods. Recent approaches have extended this paradigm through species-specific models \cite{sabathier2024animalavatar,ruegg2023bite,rueegg2022barc,li2021coarse,biggs2020wldo,biggs2018creatures,tan2023distilling,cho2025dogrecon,zuffi2024varen,li2021hsmal,li2024dessie,zuffi2019threezebra,wang2021birds,kanazawa2018CMR,badger20203d} or unified multi-species frameworks \cite{li20243dfauna,lyu2025animer,an2025animer+}. While these methods have achieved impressive progress, they remain constrained by limited 3D supervision and highly imbalanced species distributions.

We identify two key bottlenecks behind this limitation. First, existing models lack explicit morphological knowledge beyond what can be learned from annotated training data, causing reconstructions to collapse toward the average shapes of dominant species. Second, obtaining large-scale 3D supervision for animals remains expensive and difficult. Although abundant 2D animal annotations are available, converting them into reliable 3D supervision remains an open challenge. Existing approaches either rely on computationally expensive fitting pipelines \cite{xu2023animal3d,biggs2018creatures} or synthetic datasets \cite{niewiadomski2025generative,lyu2025animer,kulits2025raw,Mu_2020_CVPR,zuffi2024awol,zuffi2019threezebra}, both of which have limitations in scalability or realism.

We argue that addressing long-tail animal reconstruction requires both stronger priors and broader supervision. To this end, we introduce PRIMA, a framework that combines \textcolor{blue}{taxonomic shape priors} and adaptation-driven data bootstrapping for robust monocular animal mesh recovery (Fig.~\ref{fig:teaser}). First, we leverage BioCLIP \cite{stevens2024bioclip} \textcolor{blue}{image embeddings as taxonomic shape priors, injecting species-level knowledge} into the reconstruction process and improving shape estimation for underrepresented species. Second, we introduce an adaptation-driven bootstrapping framework that refines feed-forward mesh predictions and converts abundant 2D annotations into reliable pseudo-3D supervision. This process produces Quadruped3D, a large-scale pseudo-3D dataset that substantially expands species and pose coverage without requiring additional manual 3D annotations.
Our contributions are summarized as follows:

\begin{itemize}

\item We introduce \textcolor{blue}{taxonomic shape priors} for animal mesh recovery by leveraging BioCLIP \textcolor{blue}{image} embeddings to provide species-aware shape conditioning, improving reconstruction fidelity for underrepresented species.

\item We propose an adaptation-driven data bootstrapping framework that converts large-scale 2D animal datasets into reliable pseudo-3D supervision and enables the construction of the Quadruped3D dataset.

\item By combining \textcolor{blue}{taxonomic shape priors} with scalable pseudo-3D supervision, PRIMA establishes state-of-the-art performance across Animal3D, CtrlAni3D, Quadruped2D, and Animal Kingdom, particularly on rare species and challenging poses.

\end{itemize}

\section{Related Work}
\label{sec:relatedwork}

\subsection{Learning-based Animal Mesh Recovery}

The primary challenge in learning-based 3D animal reconstruction lies in the extreme morphological diversity across species and the inherent scarcity of 3D ground truth. Early research addressed this by developing high-fidelity regressors for well-documented species, such as BARC \cite{rueegg2022barc}, BITE \cite{ruegg2023bite}, DogRecon \cite{cho2025dogrecon}, and Animal Avatars \cite{sabathier2024animalavatar} for dogs, or specialized frameworks for horses \cite{zuffi2024varen}, zebras \cite{zuffi2019threezebra}, and birds \cite{wang2021birds, kanazawa2018CMR}. While these methods achieve impressive results within their respective domains, their reliance on category-specific priors limits their scalability to the vast ``long-tail'' of animal species found in the wild. 

To overcome this bottleneck, recent works have shifted toward unified frameworks. Some works \cite{li20243dfauna, yao2023hi, yao2022lassie, kulkarni2020articulation} learn pan-category models from image or video collections via a shared shape bank or canonical surface mapping, while AniMer \cite{lyu2025animer} utilizes family-aware transformers to handle diverse quadrupeds. AniMer+ \cite{an2025animer+} further extends this scope to birds by incorporating a Mixture-of-Experts (MoE) design to manage inter-class variance. However, existing methods remain constrained by the long-tail distribution of animal datasets, where a few dominant species bias the model, limiting generalization to sparsely observed taxa. To mitigate this, our work incorporates a \textcolor{blue}{taxonomic shape prior} derived from BioCLIP \cite{stevens2024bioclip}, \textcolor{blue}{a foundation model whose image encoder is trained under hierarchical taxonomic supervision. We use its image embeddings to provide a species-aware shape initialization} that improves robustness across the long-tail without requiring explicit per-species supervision.

\subsection{Pseudo-3D Supervision and Adaptation-Based Refinement}
The analysis-by-synthesis paradigm addresses the ambiguity of monocular reconstruction by optimizing 3D parameters to match 2D observations. Classical fitting-based approaches directly optimize the parameters of parametric models, including SMPLify \cite{bogo2016keep} and SMPLify-X \cite{pavlakos2019expressive} for humans \cite{loper2015smpl}, and SMALify \cite{biggs2018creatures} for animals. While these methods can achieve accurate alignment through strong geometric priors \cite{rempe2021humor,zioulis2023kbody}, they are computationally expensive and highly sensitive to initialization.

A closely related line of work combines feed-forward prediction with optimization-based refinement to improve supervision quality. In human mesh recovery, some works \cite{kolotouros2019learning, joo2021exemplar} leverage optimization to refine model predictions and subsequently retrain the network using improved pseudo labels. These approaches demonstrate that optimization can serve not only as an inference-time refinement tool but also as an effective mechanism for generating higher-quality supervision. Similar refinement strategies have also been explored through online adaptation and post-hoc optimization \cite{guan2022out,nam2023cyclic,bidulka2025escape,wang2026uncertainty}.

For animal reconstruction, the scarcity of large-scale 3D supervision has motivated the use of optimization-based fitting pipelines to generate pseudo-SMAL annotations \cite{biggs2018creatures,xu2023animal3d}. Recent works have further explored weakly supervised learning and synthetic data generation to alleviate annotation costs \cite{zuffi2024awol,niewiadomski2025generative,lyu2025animer,kulits2025raw}. However, existing approaches remain limited either by optimization cost, manual verification requirements, or the domain gap between synthetic and real-world data.

Our work differs from prior refinement-based approaches in that we use adaptation primarily as a data bootstrapping mechanism rather than an inference-time enhancement technique. By generating reliable pseudo-3D supervision from large-scale 2D animal datasets, our approach substantially expands the scale and diversity of available 3D supervision, enabling improved generalization without additional manual annotations.

\begin{figure*}[t]
\vspace{-5pt}
    \centering
    \includegraphics[width=1.0\linewidth]{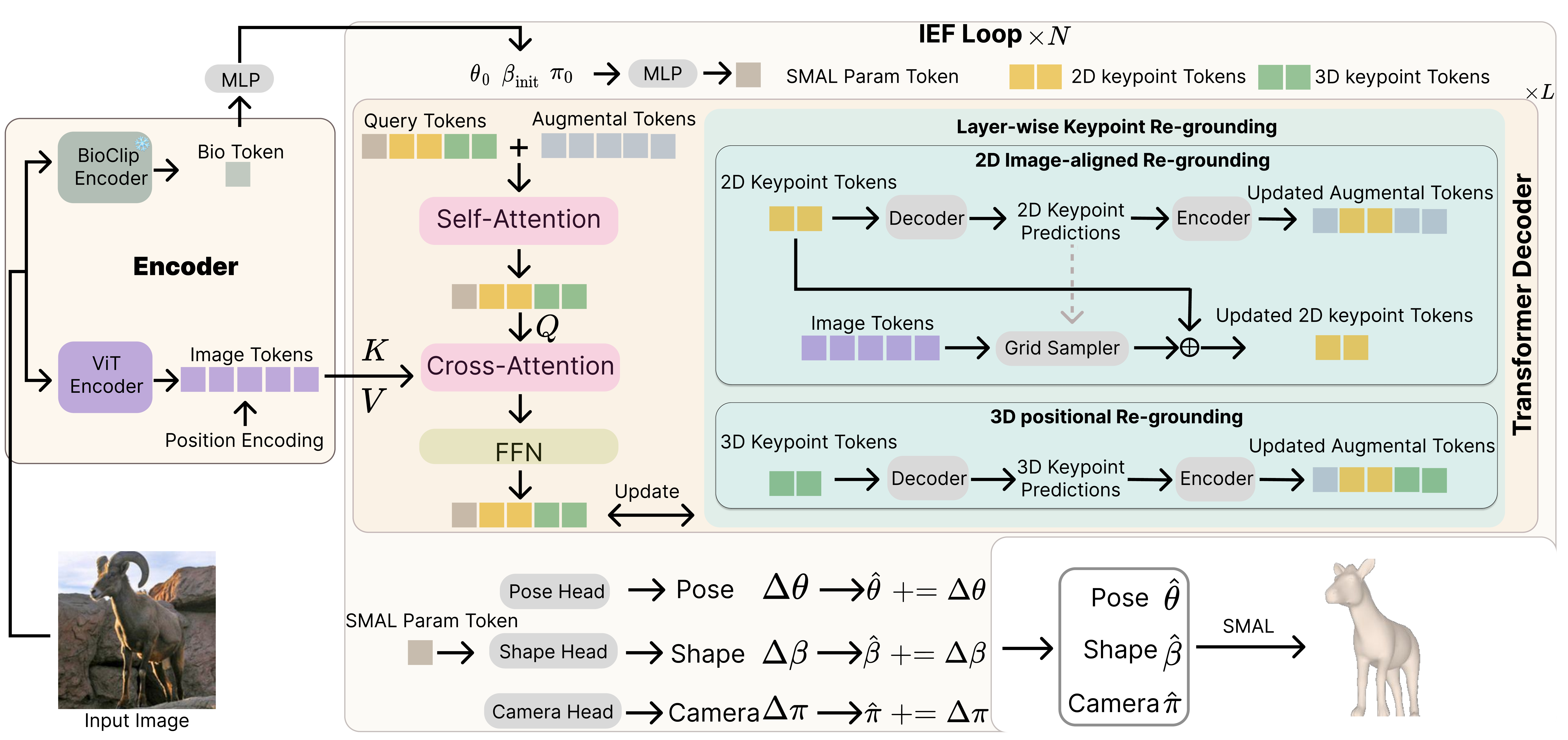}
    \caption{\textbf{Overview of the proposed model architecture.}     \textcolor{blue}{Given an input image, a ViT encoder extracts image tokens and a BioCLIP encoder produces a bio token, which an MLP maps to the initial SMAL shape parameter $\beta_{\text{init}}$. The transformer decoder and regression heads then follow an Iterative Error
    Feedback (IEF) scheme for $N$ iterations: each iteration projects the current parameters $(\hat{\theta}, \hat{\beta}, \hat{\pi})$ into the SMAL parameter token, passes the query tokens through all $L$ decoder layers, and applies the pose, shape, and camera heads once to regress $(\Delta\theta, \Delta\beta, \Delta\pi)$, which update the parameters for the next iteration. Within a layer, the query tokens plus the augmental tokens (zero-initialized at
    the start of each iteration) undergo self-attention, cross-attention to image tokens, and an FFN, followed by the Layer-wise Keypoint Re-grounding module: its 2D Image-aligned branch re-grounds the 2D keypoint tokens by grid sampling image features, while its 3D Positional branch encodes the 3D
    predictions, both producing updated augmental tokens for the next layer}. After the final iteration, $(\hat{\theta}, \hat{\beta})$ produce the final mesh
    through SMAL.}
    \label{fig:architecture}
\vspace{-8pt}
\end{figure*}

\section{Methods}

\subsection{Preliminaries}
\textbf{SMAL} \cite{smalZuffi:CVPR:2017} is a parametric model that represents 3D mesh geometry with shape parameters $\beta \in \mathbb{R}^{41}$ and pose parameters $\theta \in \mathbb{R}^{105}$, enabling diverse quadruped shapes and articulated poses. Given these parameters, SMAL defines a differentiable function that outputs a 3D mesh $m \in \mathbb{R}^{3889 \times 3}$.
To fit the SMAL model to a given animal image, SMALify \cite{biggs2018creatures} adopts an optimization-based fitting procedure that estimates the SMAL parameters by minimizing reprojection errors with respect to 2D keypoints and silhouettes.

\textbf{BioCLIP} \cite{stevens2024bioclip} is a vision foundation model for the tree of life, trained on large-scale biological data spanning diverse organisms (including a wide range of animal species) with hierarchical taxonomic labels. The image encoder maps an input image to a species-aware embedding in a shared feature space, capturing \textcolor{blue}{species-discriminative visual cues and taxonomic semantics.} 

\subsection{Model Architecture}
\label{sec:architecture}
As shown in Figs.~\ref{fig:teaser} and~\ref{fig:architecture}, our model adopts an encoder-decoder design with two parallel visual encoders. A ViT extracts image tokens, while a BioCLIP encoder produces a bio token for bio-guided shape parameter initialization. A keypoint-aware transformer decoder refines learnable keypoint and parameter tokens layer by layer, and regression heads predict residual updates to the SMAL parameters to recover the final 3D animal mesh.

\subsubsection{Image Encoders}
To capture both general appearance and species semantics, we extract features from two parallel encoders applied to the cropped and normalized animal image $I$. A standard ViT encoder $f_{\text{ViT}}$ produces general image features:
\begin{equation}
    F_{\text{img}} = f_{\text{ViT}}(I) \in \mathbb{R}^{N \times D},
\end{equation}
where $N$ denotes the number of visual tokens and $D$ denotes the feature dimension. A frozen BioCLIP encoder $f_{\text{bio}}$ produces species-aware features:
\begin{equation}
    F_{\text{bio}} = f_{\text{bio}}(I) \in \mathbb{R}^{1 \times 512}.
\end{equation}
The image feature $F_{\text{img}}$ captures instance-specific visual details, whereas $F_{\text{bio}}$ encodes \textcolor{blue}{taxonomic priors that reflect species-level semantics.}
To leverage \textcolor{blue}{this taxonomic knowledge} and alleviate the challenges posed by the unbalanced category distributions, we introduce a \textcolor{blue}{taxonomy-informed} coarse-to-fine shape initialization strategy. 
Instead of initializing SMAL shape parameters from a zero vector or a generic mean template, we first derive a taxonomic shape prior from the species-aware features, which provides a semantically meaningful initialization in the SMAL shape space. 
Subsequent geometric refinement is then performed by the transformer decoder,
\textcolor{blue}{which predicts an instance-specific shape residual} conditioned
on image features.

Concretely, the species-aware feature is projected into a shared latent space to form a bio-token $T_{\text{bio}}$. 
The initial SMAL shape parameter is predicted as:
\begin{equation}
    \beta_{\text{init}} = \mathcal{M}(T_{\text{bio}}) \in \mathbb{R}^{1 \times 41},
\end{equation}
where $\mathcal{M}$ is a multi-layer perceptron (MLP) that maps the \textcolor{blue}{taxonomic prior} to the SMAL shape parameter space. 
\textcolor{blue}{
Importantly, $\beta_{\text{init}}$ serves only as a coarse shape prior rather
than the final shape prediction. The decoder further predicts an
instance-specific residual $\Delta\beta$ from the image features, yielding
the final shape parameter as
}
\begin{equation}
    \textcolor{blue}{\beta = \beta_{\text{init}} + \Delta\beta.}
\end{equation}

\subsubsection{Keypoint-Aware Decoder}
Inspired by the keypoint tokens in 3DB \cite{yang2026sam3dbody} for human pose decoding, we propose a Keypoint-Aware SMAL Decoder, a transformer-based iterative refinement module that integrates structured token reasoning for animal mesh recovery. 
Unlike previous methods \cite{lyu2025animer} that directly regress pose and shape parameters from global features via regression heads, our decoder explicitly models SMAL parameter token and keypoint tokens.

The complete set of query tokens is defined as:
\begin{equation}
    T = [T_\text{param}, T_\text{2Dkeypoint}, T_\text{3Dkeypoint}],
\end{equation}
where $T_\text{param}$, $T_\text{2Dkeypoint}$, and $T_\text{3Dkeypoint}$ denote the SMAL parameter, 2D keypoint, and 3D keypoint query tokens, respectively. 
Each decoder layer applies self-attention over the query tokens,
cross-attention with the image features $F_\text{img}$, and a
feed-forward network.

\textcolor{blue}{
To facilitate progressive geometric reasoning, we introduce a
\textbf{Layer-wise Keypoint Re-grounding} module comprising
\textbf{2D Image-aligned Re-grounding} and
\textbf{3D Positional Re-grounding} branches
(Fig.~\ref{fig:architecture}).
After each decoder layer except the last, the 2D and 3D keypoint tokens
are decoded into their corresponding keypoint predictions.
The 2D predictions are re-encoded into augmental tokens and guide
differentiable grid sampling of image-aligned features, which are
injected into the 2D keypoint tokens.
The 3D predictions are re-encoded only into positional augmental tokens.
Both the decoding and re-encoding steps use a single pair of lightweight MLPs
shared across layers and IEF iterations, adding negligible parameters.
}
After the stacked decoder layers, the decoder produces:
\begin{equation}
    O = \mathcal{D}(T, F_\text{img}) \in \mathbb{R}^{(1 + 2J) \times D},
\end{equation}
where $\mathcal{D}$ consists of stacked self-attention, cross-attention, and feed-forward layers, and $J$ denotes the number of joints. 
Through cross-attention, the decoder integrates keypoint-aware representations with visual context, enabling structured shape and pose estimation.

The first output token of $O$ is fed into regression heads to predict residual
pose, shape, and camera parameters.
\textcolor{blue}{In particular, the predicted shape residual $\Delta\beta$
refines the BioCLIP-based coarse initialization $\beta_{\text{init}}$ to
obtain the final instance-specific shape $\beta$.}

\subsection{Model Training}
Our model is trained with a multi-task loss term:
\begin{equation}
    \mathcal{L} =
\lambda_\text{param} \mathcal{L}_{\text{param}}
+
\lambda_\text{keypoint} \mathcal{L}_{\text{keypoint}}.
\end{equation}

\textbf{Parameter Losses.} For 3D data with SMAL parameters, we apply $L_2$ regression losses on the parameters and prior losses to encourage plausible poses.  

\textbf{2D/3D Keypoint Loss.}
Keypoint losses consist of two terms: the auxiliary 2D and 3D keypoints predicted by the decoder, which provide deep supervision, and the 2D and 3D keypoints regressed from the mesh, which ensure reprojection correctness. Both use an $L_1$ loss.

\subsection{Adaptation-Driven Data Bootstrapping}

\subsubsection{Adaptation for Pseudo-3D Supervision}

A major obstacle to animal mesh recovery is the lack of large-scale 3D supervision. While abundant 2D animal annotations are available, directly obtaining reliable 3D labels remains expensive and difficult. To address this challenge, we introduce an adaptation-based refinement procedure that converts feed-forward mesh predictions into high-quality pseudo-3D supervision.
\begin{figure}[h]
    \centering
    \includegraphics[width=\linewidth]{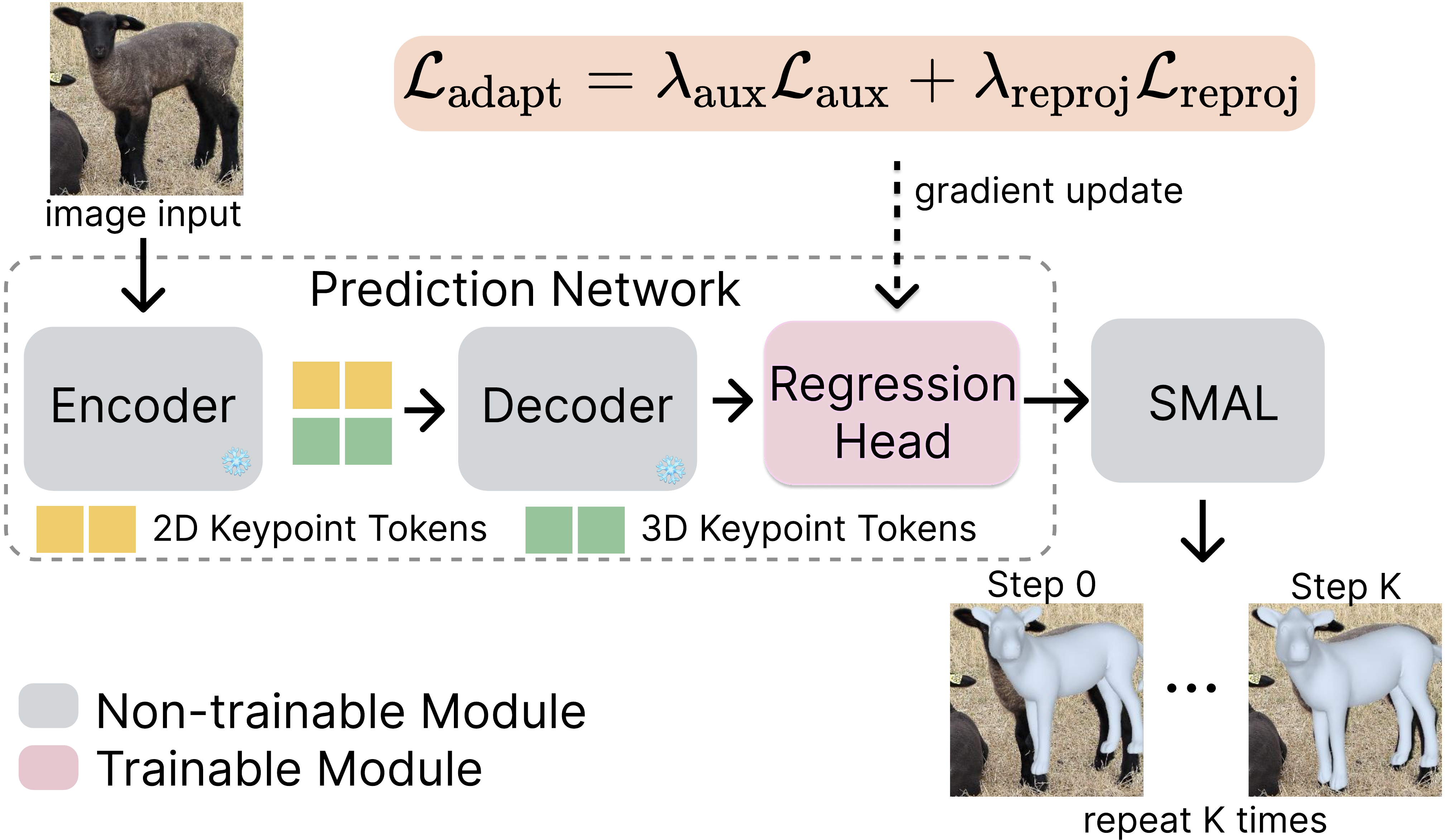}
    \caption{\textbf{Adaptation framework.}}
    \label{fig:adaptation}
    \vspace{-10pt}
\end{figure}
Given a pre-trained PRIMA model, we refine the predicted SMAL parameters using available 2D keypoint annotations. \textcolor{blue}{As PRIMA is pre-trained with joint 2D/3D keypoint supervision, this refinement starts from a 3D-consistent initialization and largely preserves 3D structure for a moderate number of steps.} Notably, this adaptation procedure is designed as a data bootstrapping mechanism for pseudo-label generation rather than a component of the final evaluation pipeline.
As shown in Fig.~\ref{fig:adaptation}, we freeze the image encoder and transformer decoder and optimize only the SMAL regression head together with the 2D/3D keypoint tokens. This design preserves the learned geometric prior while allowing instance-specific refinement of shape and pose.
The optimization objective is defined as:
\begin{equation}
\mathcal{L}_{\text{adapt}} =
\lambda_\text{aux} \mathcal{L}_{\text{aux}}
+
\lambda_\text{reproj} \mathcal{L}_{\text{reproj}},
\end{equation}
where $\mathcal{L}_{\text{aux}}$ enforces consistency between the predicted auxiliary keypoints and the annotated 2D keypoints, and $\mathcal{L}_{\text{reproj}}$ minimizes the reprojection error between annotated 2D keypoints and projected 3D keypoints.
By restricting optimization to a small subset of parameters and performing only a limited number of update steps, the refinement process remains efficient while substantially improving pose alignment and articulation accuracy. \textcolor{blue}{We further provide direct 3D validation of the adaptation process against Animal3D ground truth in Supp.~Sec.~D.}

\subsubsection{Quadruped3D Construction}

The proposed adaptation procedure enables us to transform large-scale 2D animal annotations into pseudo-3D supervision. \textcolor{blue}{We apply it to the 14K Quadruped2D
\cite{ye2024superanimal} training images (with the remaining 10K reserved
for testing). Samples with fewer than 13 visible keypoints are excluded,
and adapted samples with low final reprojection quality are further
filtered, retaining about 9K samples. The effects of keypoint visibility
and the 3D validity of the same 2D-guided adaptation are analyzed in
Supp.~Sec.~D.}

Applying the proposed bootstrapping pipeline produces about 9K high-quality pseudo-3D annotations across 24 quadruped species, forming the Quadruped3D dataset. Compared with Animal3D \cite{xu2023animal3d}, which has 3K pGT SMAL annotations, Quadruped3D adds three times as much as Animal3D and greatly increases species and pose diversity, all without any manual 3D annotation. The dataset is available at \href{https://zenodo.org/records/23662465}{https://zenodo.org/records/23662465}.
Fig.~\ref{fig:quadruped3d} presents representative examples from Quadruped3D. More qualitative results are in the Supplementary.
The generated pseudo ground-truth (pGT) is subsequently merged with the original training set and used to retrain our proposed mesh recovery model (Sec.~\ref{sec:architecture}). 

\begin{figure}[h]
    \centering
    \includegraphics[width=0.9\linewidth]{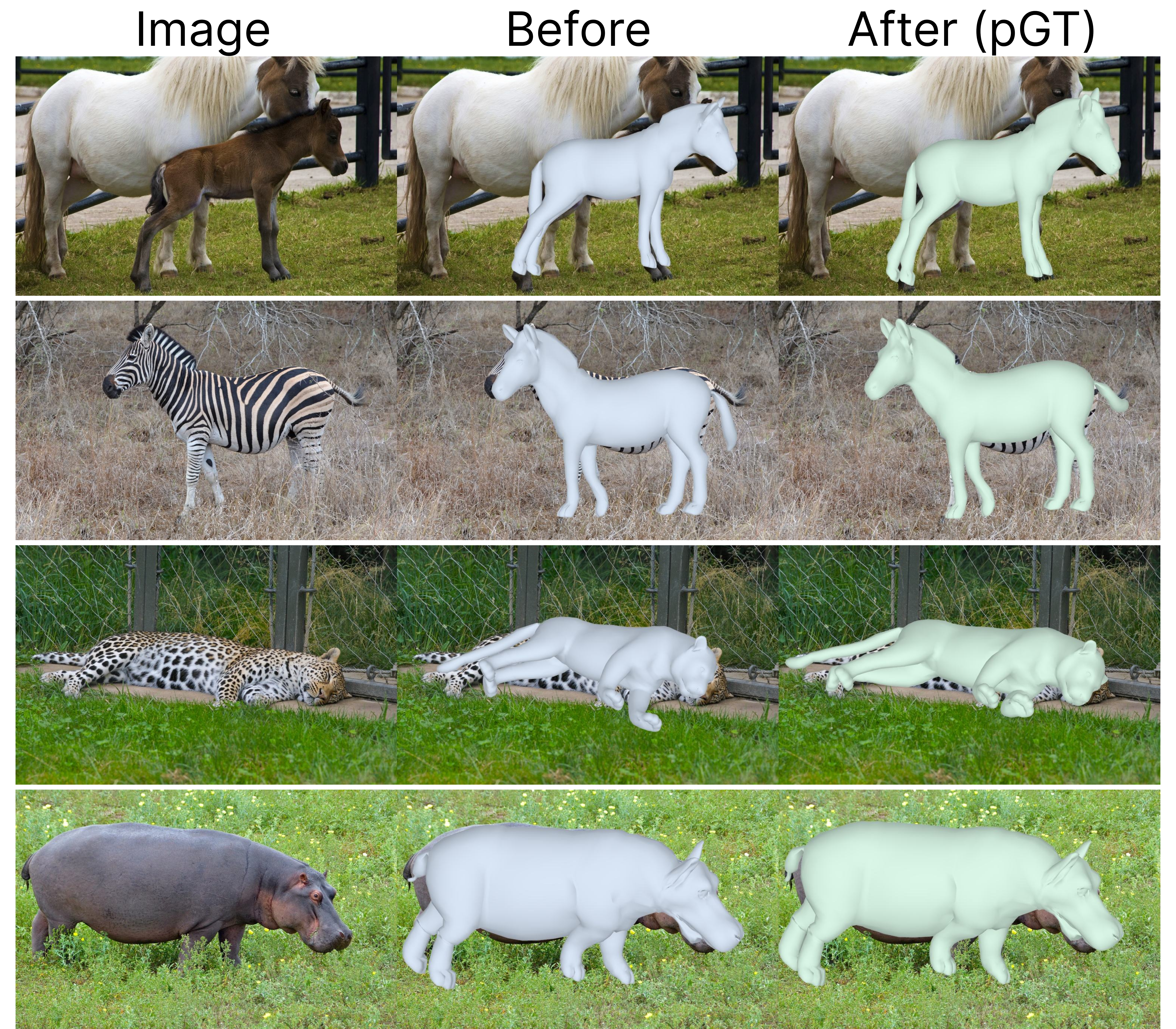}
    \caption{\textbf{Quadruped3D pseudo-3D generation examples.} 
Each row shows the input image (left), the initial feed-forward prediction (middle), and the refined pseudo-3D annotation used for data bootstrapping (right). The adaptation procedure improves pose alignment and shape fidelity across diverse species and articulated poses, enabling reliable pseudo-3D supervision from 2D animal annotations.}
    \label{fig:quadruped3d}
    \vspace{-10pt}
\end{figure}

\section{Experiments}
\label{sec:experiments}

\subsection{Datasets}
PRIMA is trained using a combination of real-image 3D annotations, synthetic data, and large-scale 2D animal supervision.
For the first round of training, we jointly use Animal3D \cite{xu2023animal3d}, CtrlAni3D \cite{lyu2025animer}, and Quadruped2D \cite{ye2024superanimal}. Animal3D provides 3K real-image animal instances with pGT SMAL annotations, while CtrlAni3D is a large-scale synthetic dataset generated for multi-species animal reconstruction. Quadruped2D is a subset of Quadruped80K containing 24K images with 2D keypoint annotations.

We further incorporate our proposed Quadruped3D dataset, which provides pseudo-3D SMAL annotations for a subset of Quadruped2D.
\textcolor{blue}{The final training stage uses the same three datasets, with the 9K Quadruped2D images covered by Quadruped3D additionally supervised in 3D.}

Following prior work \cite{lyu2025animer,an2025animer+}, we also evaluate out-of-distribution generalization on Animal Kingdom \cite{ng2022animalkingdom}, which features diverse species and challenging in-the-wild poses.

\begin{figure*}
    \centering
    \includegraphics[width=0.85\linewidth]{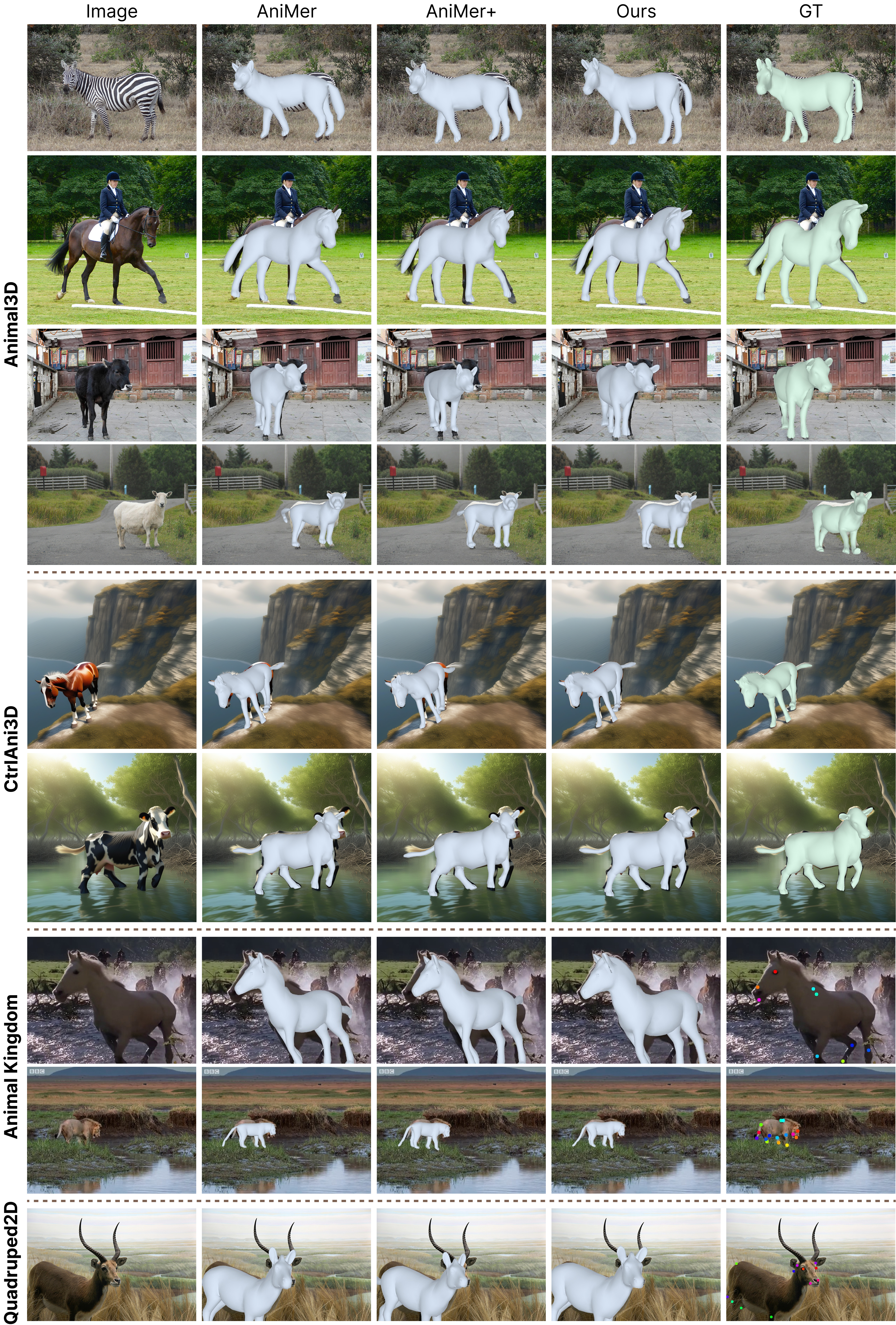}
    \caption{\textbf{Qualitative comparisons on Animal3D (rows 1--4), CtrlAni3D (rows 5--6), Animal Kingdom (rows 7--8), and Quadruped2D (row 9) datasets.} We compare our results with AniMer \cite{lyu2025animer} and AniMer+ \cite{an2025animer+}.}
    \label{fig:qualitative}
\end{figure*}

\subsection{Metrics}
We evaluate model performance using both 3D and 2D metrics. For 3D evaluation, we report PA-MPJPE (hereafter abbreviated as PAJ), defined as the Procrustes-aligned mean per-joint position error, and PA-MPVPE (hereafter abbreviated as PAV), defined as the Procrustes-aligned mean per-vertex position error computed over the SMAL mesh vertices.
For 2D evaluation, we use PCK (Percentage of Correct Keypoints), measuring the proportion of predicted keypoints within a given threshold of the ground truth. We report PCK@0.1 and PCK@0.15 (denoted P@0.1 and P@0.15 in tables). Additionally, we report AUC (Area Under the Curve), obtained by integrating PCK over thresholds from 0 to 1.
Notably, both the 2D and 3D keypoints used for metric computation are derived from the predicted mesh.
To evaluate whether the predicted animal shapes capture meaningful family-level structure, we compute the Adjusted Rand Index (ARI) between the low-dimensional representation of the shape parameters and ground-truth family labels. ARI measures similarity between two partitions while correcting for chance, ranging from -1 to 1, where 1 indicates perfect alignment, and 0 corresponds to random assignment. \textcolor{blue}{ARI is averaged over 10 $k$-means seeds (0--9, \texttt{n\_init}$=$10); the standard deviation is zero unless stated.} 

\subsection{Comparison with State-of-the-Art Methods}
Following the evaluation protocols of Animal3D \cite{xu2023animal3d} and AniMer+ \cite{an2025animer+}, we quantitatively compare our model with several baselines, including HMR \cite{kanazawa2018hmr}, HMR2.0 \cite{goel2023hmr2}, GenZoo \cite{niewiadomski2025generative}, WLDO \cite{biggs2020wldo}, AniMer \cite{lyu2025animer}, and AniMer+ \cite{an2025animer+}. Since GenZoo is trained with the SMAL+ \cite{zuffi2024awol} model, we limit our evaluation to 26 3D joints and 2D metrics. For qualitative comparisons, we focus on AniMer and AniMer+, which also tackle multi-species mesh recovery.
As shown in Tab.~\ref{tab:quan_3D}, our model achieves state-of-the-art performance on two 3D datasets across both 3D and 2D evaluation metrics. Furthermore, as reported in Tab.~\ref{tab:quan_2d}, our approach maintains SOTA performance on Quadruped2D and the challenging out-of-domain Animal Kingdom test set. \textcolor{blue}{AniMer and AniMer+ are trained on 41.3K samples (24.2K with 3D and 17.1K with 2D supervision). PRIMA uses 25K samples (11K 3D and 14K Q2D), with 9K of the Q2D samples further augmented with pseudo-3D supervision.}

The qualitative results in Fig.~\ref{fig:qualitative} further highlight the advantages of our approach. Overall, the meshes predicted by our model exhibit better alignment across major body parts, including the legs, tail, and head. Moreover, our method more accurately preserves the characteristic shape of the correct animal family compared to the baselines.
For example, in the first row, AniMer and AniMer+ predict a dog-like shape given a zebra input. Similarly, in the fourth row (sheep), both baseline methods produce predictions resembling Felidae-like shapes, whereas our model successfully captures the distinctive sheep-like shape.
Additional qualitative results are provided in the Supplementary Material.

\begin{table}[h]
\centering
\scriptsize
\setlength{\tabcolsep}{4pt}
\renewcommand{\arraystretch}{1.1}
\begin{tabular}{c|ccc|ccc}
\toprule
 & \multicolumn{3}{c|}{Animal3D} 
 & \multicolumn{3}{c}{CtrlAni3D} \\
Method 
 & AUC$\uparrow$  & PAJ$\downarrow$ & PAV$\downarrow$
 & AUC$\uparrow$  & PAJ$\downarrow$ & PAV$\downarrow$ \\
\midrule
HMR$^\dagger$ \cite{kanazawa2018hmr}    & 76.2  & 124.1 & 136.0 & 78.5  & 135.2 & 148.0 \\
WLDO \cite{biggs2020wldo}      & 78.2  & 112.3 & 125.2 & 88.7  & 71.5 & 83.4 \\
HMR2.0 \cite{goel2023hmr2}    & 86.7  & 94.1 & 98.5 & 91.8  & 60.9 & 66.4 \\
GenZoo$^*$ \cite{niewiadomski2025generative} & 86.6 & 138.5 & -- & 92.9 &116.4 & -- \\
AniMer \cite{lyu2025animer}      & 88.9  & 80.4 & 85.7 & 93.8  & 44.1 & 47.6 \\
AniMer+ \cite{an2025animer+}     & 88.9 & 78.0 & 82.7 & 93.6  & 47.8 & 50.8 \\
\midrule

Ours              & \textbf{89.8}  & \textbf{75.0} & \textbf{79.2} & \textbf{94.1} & \textbf{40.6}   & \textbf{43.4}  \\
\bottomrule 
\end{tabular}
\vspace{1mm}
\\ \footnotesize $^*$ GenZoo's PAJ is computed on 26 joints. $^\dagger$ Re-evaluated by us.
\caption{Quantitative comparison on 3D datasets: Animal3D and CtrlAni3D. \textbf{Bold}: best.}
\label{tab:quan_3D}
\end{table}

\begin{table}[h]
\centering
\scriptsize
\setlength{\tabcolsep}{1.6pt}\renewcommand{\arraystretch}{1.1}
\begin{tabular}{c|ccc|ccc}\toprule
 & \multicolumn{3}{c|}{Quadruped2D} 
 & \multicolumn{3}{c}{Animal Kingdom}  \\
 Method 
 & AUC$\uparrow$ & P@0.1$\uparrow$ & P@0.15$\uparrow$
 & AUC$\uparrow$ & P@0.1$\uparrow$ & P@0.15$\uparrow$ \\
 \midrule
AniMer \cite{lyu2025animer} & 87.6 & 63.8 & 78.6 & 82.9 & 34.9 & 54.7  \\
AniMer+ \cite{an2025animer+} & 86.7 & 60.7 & 76.8 & 82.6 & 34.2 & 53.0  \\
\midrule

Ours & \textbf{93.0} & \textbf{83.3} & \textbf{91.3} & \textbf{83.6} & \textbf{38.2} & \textbf{56.7}  \\
\bottomrule
\end{tabular}
\caption{Quantitative comparison on 2D datasets: Quadruped2D and Animal Kingdom. \textbf{Bold}: best.}
\label{tab:quan_2d}
\end{table}

\begin{figure}[t]
    \centering
    \includegraphics[width=\linewidth]{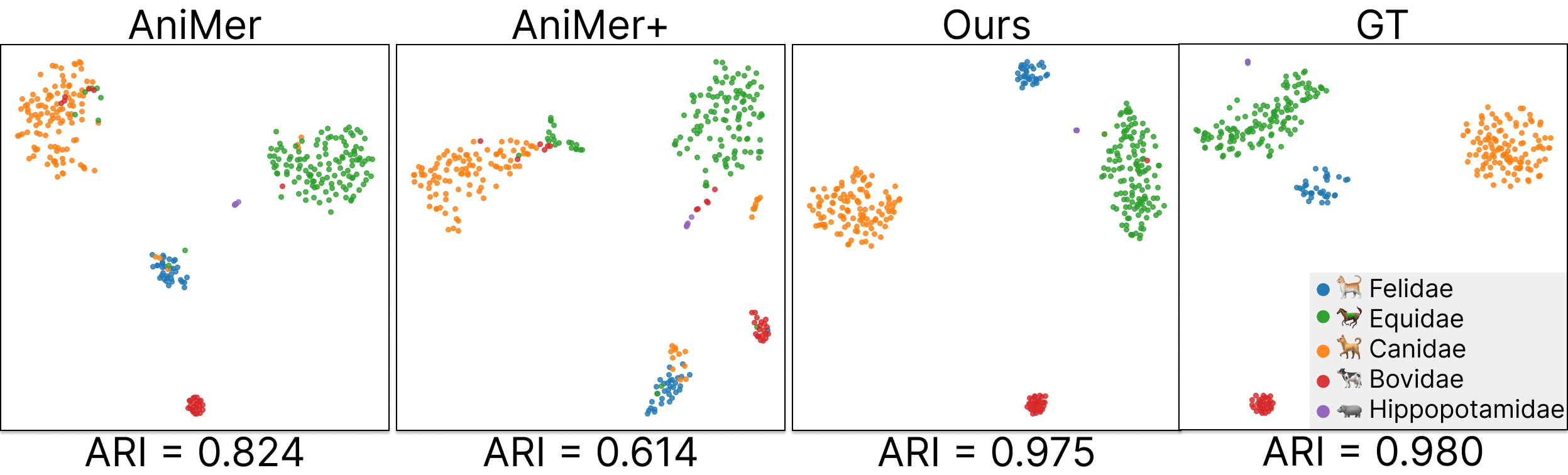}
    \caption{
    \textbf{t-SNE \cite{van2008tsne} embeddings of predicted and GT shape parameters \textcolor{blue}{$\beta$} on Animal3D.} Colors indicate different animal family category labels.}
    \vspace{-10pt}
    \label{fig:tsne_beta}
\end{figure}

\textcolor{blue}{To analyze shape prediction, we use t-SNE \cite{van2008tsne} to project the predicted and GT shape parameters $\beta \in \mathbb{R}^{41}$ into a 2D embedding on the Animal3D test set.}
As shown in Fig.~\ref{fig:tsne_beta}, our method produces compact and well-separated clusters that closely resemble the GT distribution. In contrast, AniMer \cite{lyu2025animer} and AniMer+ \cite{an2025animer+} exhibit more dispersed embeddings with noticeable cluster overlap and outliers. This observation is consistent with the ARI scores. Our method achieves an ARI of \textcolor{blue}{0.975}, approaching the GT clustering score (\textcolor{blue}{$0.980\pm0.011$}), and significantly outperforming AniMer (\textcolor{blue}{0.824}) and AniMer+ (\textcolor{blue}{0.614}), indicating substantially better alignment with the underlying shape categories.

\subsection{Ablation Study}
We perform ablation studies to evaluate the key components of PRIMA: model architecture, \textcolor{blue}{taxonomic shape priors}, and adaptation-driven data bootstrapping. 

\subsubsection{Ablation on model architecture}
We evaluate the two key architectural components: the initialized shape prior and the learnable keypoint tokens.

\textbf{w/o $\beta_{\text{init}}$.}
We remove the BioCLIP-based shape initialization and instead initialize the SMAL shape coefficients to zero, eliminating the proposed coarse-to-fine strategy.

\textbf{w/o $T_{\text{keypoint}}$.}
We remove the learnable keypoint tokens and directly regress SMAL parameters from the global SMAL token, without keypoint-aware geometric reasoning.

\begin{figure}[h]
    \centering
    \includegraphics[width=0.98\linewidth]{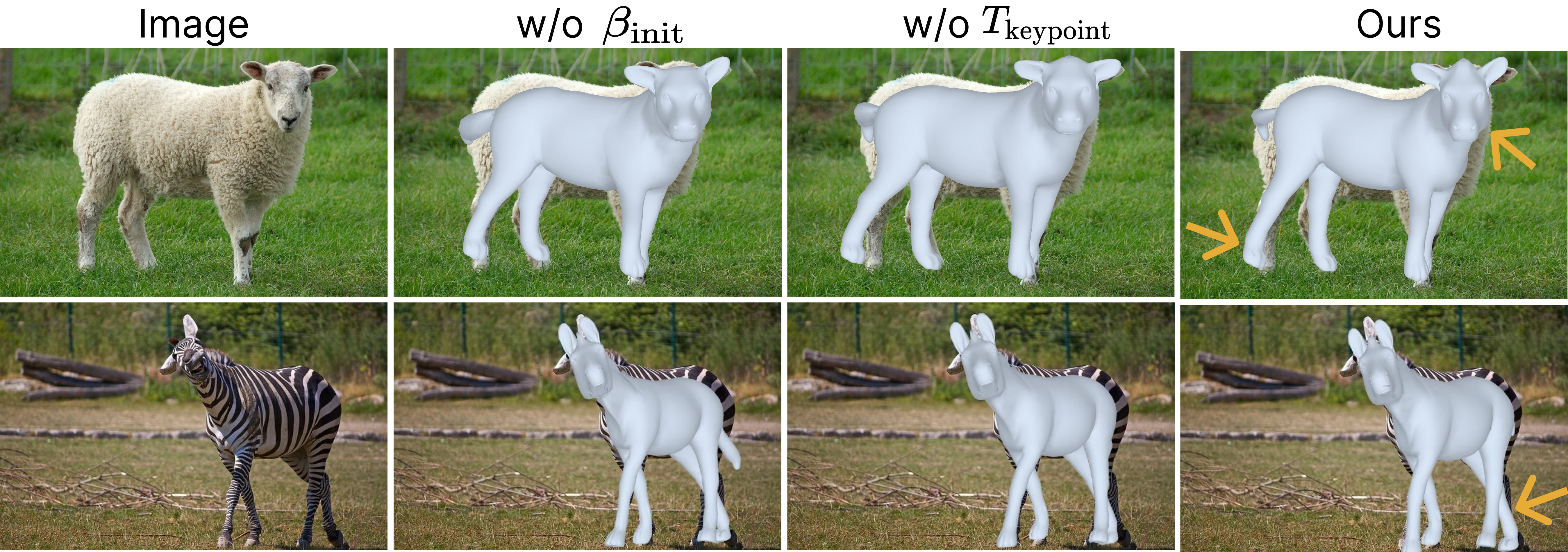}
    \caption{\textbf{Ablation on model architecture.} 
    }
    \vspace{-10pt}
    \label{fig:ablation}
    \vspace{-2pt}
\end{figure}

\begin{table}[h]
\centering
\scriptsize
\setlength{\tabcolsep}{4pt}
\renewcommand{\arraystretch}{1.1}
\begin{tabular}{c|ccc|cc|cc}
\toprule
 & \multicolumn{3}{c|}{Animal3D} 
 & \multicolumn{2}{c|}{CtrlAni3D} 
 & \multicolumn{2}{c}{Quadruped2D} \\
Method 
 & PAJ$\downarrow$ & PAV$\downarrow$ & \textcolor{blue}{ARI$\uparrow$}
 & PAJ$\downarrow$ & PAV$\downarrow$ 
  & AUC$\uparrow$ & P@0.1$\uparrow$  \\
  
\midrule
w/o $\beta_{\text{init}}$, $T_{\text{keypoint}}$ &78.2 & 82.0 & -- &45.1 &49.7 & 90.2 & 74.9  \\
w/o $\beta_{\text{init}}$   & 76.6  & 81.7  & \textcolor{blue}{0.85} & 41.8 & 44.4 & 90.9  & 78.9  \\
w/o $T_{\text{keypoint}}$    &75.6  & 80.2 & -- & 43.5 & 46.2 & 91.1 & 75.9  \\
Ours    & \textbf{75.0} & \textbf{79.2} & \textbf{\textcolor{blue}{0.98}} & \textbf{40.6} & \textbf{43.4} & \textbf{93.0} & \textbf{83.3}  \\
\bottomrule
\end{tabular}
\\ \footnotesize $^*$ ARI specifically evaluates the effect of $\beta_{\text{init}}$.
\vspace{-2mm}
\caption{\textbf{Ablation on model architecture.} }
\label{tab:ablation}
\vspace{-2mm}
\end{table}

As shown in Fig.~\ref{fig:ablation} and Tab.~\ref{tab:ablation}, both the proposed \textcolor{blue}{taxonomic shape initialization} and learnable keypoint tokens contribute to reconstruction performance. Removing either component degrades both 2D reprojection accuracy and 3D mesh quality, while the full model produces more accurate species-specific shapes and articulated poses.


\subsubsection{\texorpdfstring{Ablation on \textcolor{blue}{the taxonomic shape prior}}{Ablation on the taxonomic shape prior}}

To isolate the contribution of the proposed \textcolor{blue}{taxonomic shape prior}, we compare BioCLIP against commonly used visual representations, including CLIP \cite{radford2021learning} and DINOv2 \cite{oquab2023dinov2} features, under the same reconstruction architecture and training protocol.
As shown in Tab.~\ref{tab:prior_ablation}, BioCLIP provides consistent improvements over CLIP and DINOv2. The gains are more pronounced on underrepresented families, indicating that \textcolor{blue}{taxonomically supervised representations} provide more informative shape conditioning under long-tailed species distributions.

\begin{table}[h]
\centering
\scriptsize
\setlength{\tabcolsep}{4pt}
\renewcommand{\arraystretch}{1.1}

\begin{tabular}{c|cc|cc}
\toprule
& \multicolumn{2}{c|}{Overall}
& \multicolumn{2}{c}{Underrepresented Families} \\
\cmidrule(lr){2-3}
\cmidrule(lr){4-5}

Prior
& PAJ (A3D)$\downarrow$
& AUC (Q2D)$\uparrow$
& PAJ (A3D)$\downarrow$
& AUC (Q2D)$\uparrow$ \\
\midrule

DINOv2
& 75.8
& 91.7
& 96.4
& 86.5
 \\

CLIP
& 76.3
& 89.6
& 97.0
& 84.8
 \\

BioCLIP (Ours)
& \textbf{75.0}
& \textbf{93.0}
& \textbf{94.1}
& \textbf{90.3}
 \\

\bottomrule
\end{tabular}

\caption{\textbf{Ablation on \textcolor{blue}{the taxonomic shape prior}.}
We compare different visual priors under the same architecture and training protocol.
PAJ is evaluated on Animal3D (A3D), while AUC is evaluated on Quadruped2D (Q2D).
Underrepresented families correspond to the bottom 30\% of the training frequency distribution, including Felidae, Bovidae, and Hippopotamidae.}
\vspace{-10pt}
\label{tab:prior_ablation}

\end{table}

\subsubsection{Ablation on training with Quadruped3D}

\textcolor{blue}{
To disentangle the contributions of additional 2D data and pseudo-3D
supervision, we compare three training settings in
Tab.~\ref{tab:abla_q3d}: (i) Animal3D and CtrlAni3D only,
(ii) additionally using the 14K Q2D training images with 2D supervision,
and (iii) further providing pseudo-3D supervision to 9K eligible Q2D
samples through Q3D, denoted as Q3D Sup. The last two settings use the same 14K images,
isolating the contribution of pseudo-3D supervision from additional
real-image exposure.
}

\begin{figure}[h]
    \centering
    \includegraphics[width=\linewidth]{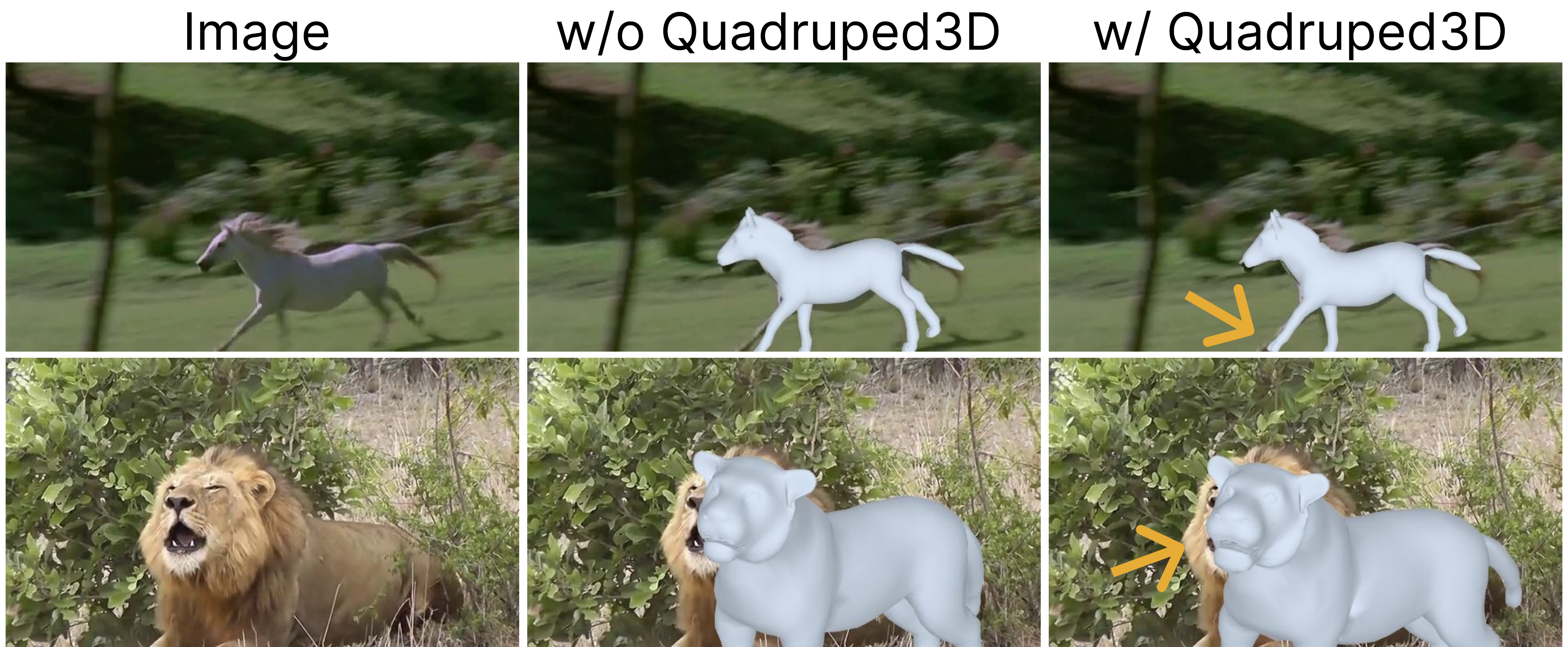}
    \caption{\textbf{Ablation on training with Quadruped3D.}}
    \vspace{-10pt}
    \label{fig:abla_q3d}
\end{figure}

\textcolor{blue}{
Adding Q2D substantially improves 2D alignment and OOD generalization
while maintaining strong 3D performance on Animal3D. Q3D further
improves all three benchmarks, showing that pseudo-3D supervision
provides additional gains without degrading 3D reconstruction.
Qualitatively, Fig.~\ref{fig:abla_q3d} further shows improved limb
articulation and mesh alignment with Q3D supervision.
}

\begin{table}[h]
\centering
\scriptsize
\setlength{\tabcolsep}{5pt}
\renewcommand{\arraystretch}{1.1}

\begin{tabular}{c|ccc|c|cc}
\toprule
& \multicolumn{3}{c|}{Training Data}
& \multicolumn{1}{c|}{PAJ$\downarrow$ }
& \multicolumn{2}{c}{P@0.1$\uparrow$ } 
 \\

& A3D+C3D & Q2D & Q3D Sup.
& A3D
& Q2D
& AK \\

\midrule

\textcolor{blue}{(i)} & \textcolor{blue}{\checkmark} & &
& \textcolor{blue}{75.5}  & \textcolor{blue}{60.6} & \textcolor{blue}{31.5}\\

\textcolor{blue}{(ii)} & \checkmark & \checkmark &
& 75.3  & 79.8  & \textcolor{blue}{37.7}\\

\textcolor{blue}{(iii)} & \checkmark & \checkmark & \checkmark
& \textbf{75.0}  & \textbf{83.3}
 & \textcolor{blue}{\textbf{38.2}} 
\\

\bottomrule
\end{tabular}

\caption{\textcolor{blue}{
\textbf{Ablation on Q2D and Q3D supervision.}
The last two settings use the same 14K Q2D images, with Q3D additionally
providing pseudo-3D supervision to 9K eligible samples. AK denotes Animal Kingdom.
}}
\label{tab:abla_q3d}
\end{table}

\section{Conclusions}
We presented PRIMA, a framework for robust monocular animal mesh recovery that addresses two fundamental challenges in long-tail animal reconstruction: limited species-aware priors and scarce 3D supervision. PRIMA combines BioCLIP-derived \textcolor{blue}{taxonomic shape priors} with adaptation-driven data bootstrapping to improve shape and pose estimation across diverse quadruped species and challenging articulations. Furthermore, the proposed bootstrapping framework enables the construction of Quadruped3D, a large-scale pseudo-3D dataset that triples the amount of available real-image 3D animal supervision without requiring additional manual annotations. Extensive experiments demonstrate that the proposed \textcolor{blue}{taxonomic shape priors} and scalable pseudo-3D supervision provide complementary benefits, establishing state-of-the-art performance across multiple benchmarks. We hope PRIMA and Quadruped3D will facilitate future research toward scalable and generalizable animal mesh recovery.

\clearpage
\onecolumn

\setcounter{figure}{0}
\setcounter{table}{0}

\renewcommand{\thefigure}{\Alph{figure}}
\renewcommand{\thetable}{\Alph{table}}


\renewcommand{\theequation}{\Alph{equation}}
\renewcommand{\thesection}{\Alph{section}}
\renewcommand{\thesubsection}{\Alph{section}.\arabic{subsection}}


\setcounter{page}{1}
\appendix
\setcounter{figure}{0}
\setcounter{table}{0}
\maketitle

\section{Dataset Statistics: Taxonomic Distribution}

\begin{figure}[htbp]
    \centering
    \begin{subfigure}[b]{0.48\linewidth}
        \centering
        \includegraphics[width=\linewidth]{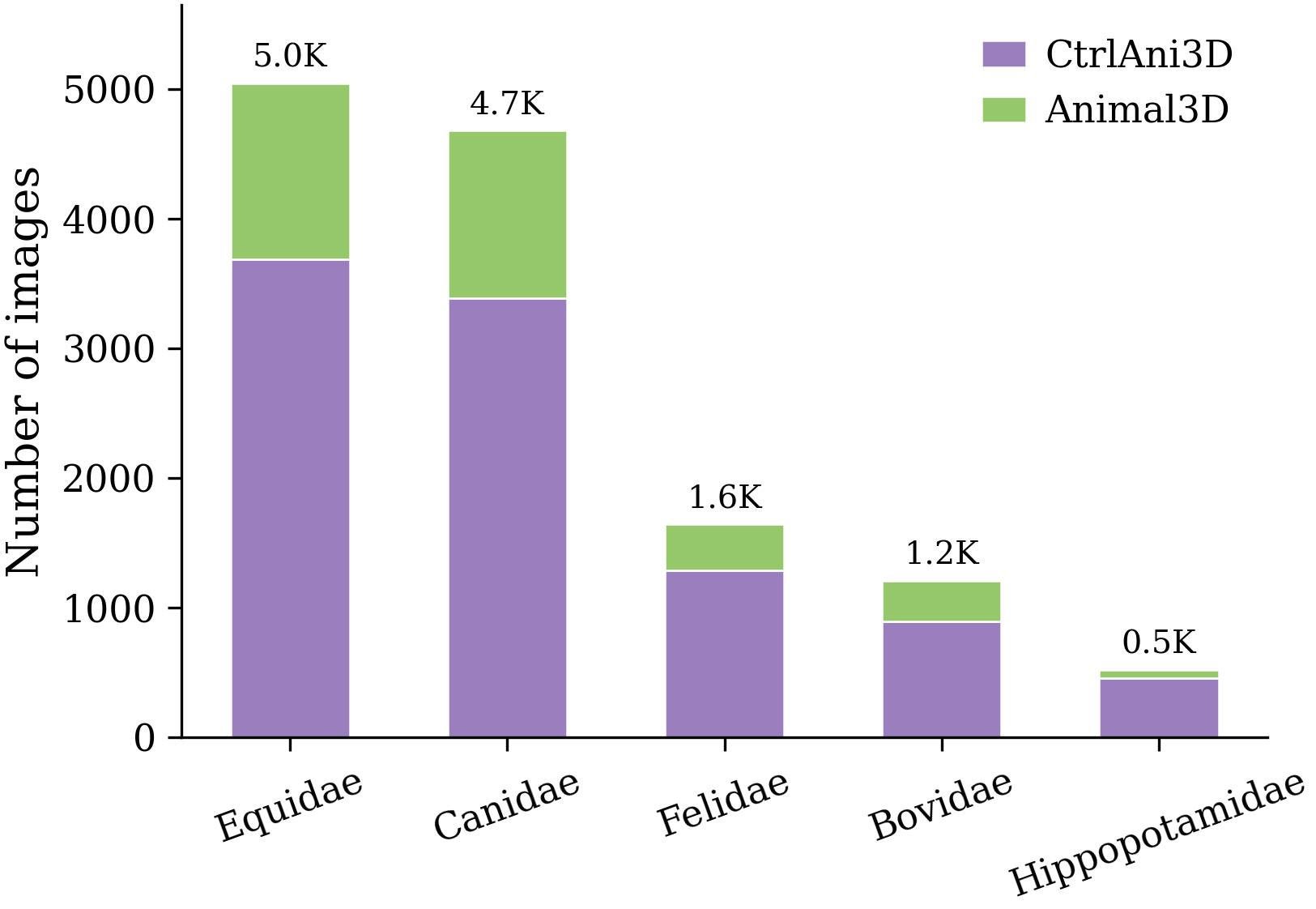}
        \caption{Animal3D and CtrlAni3D.}
    \end{subfigure}\hfill
    \begin{subfigure}[b]{0.48\linewidth}
        \centering
        \includegraphics[width=\linewidth]{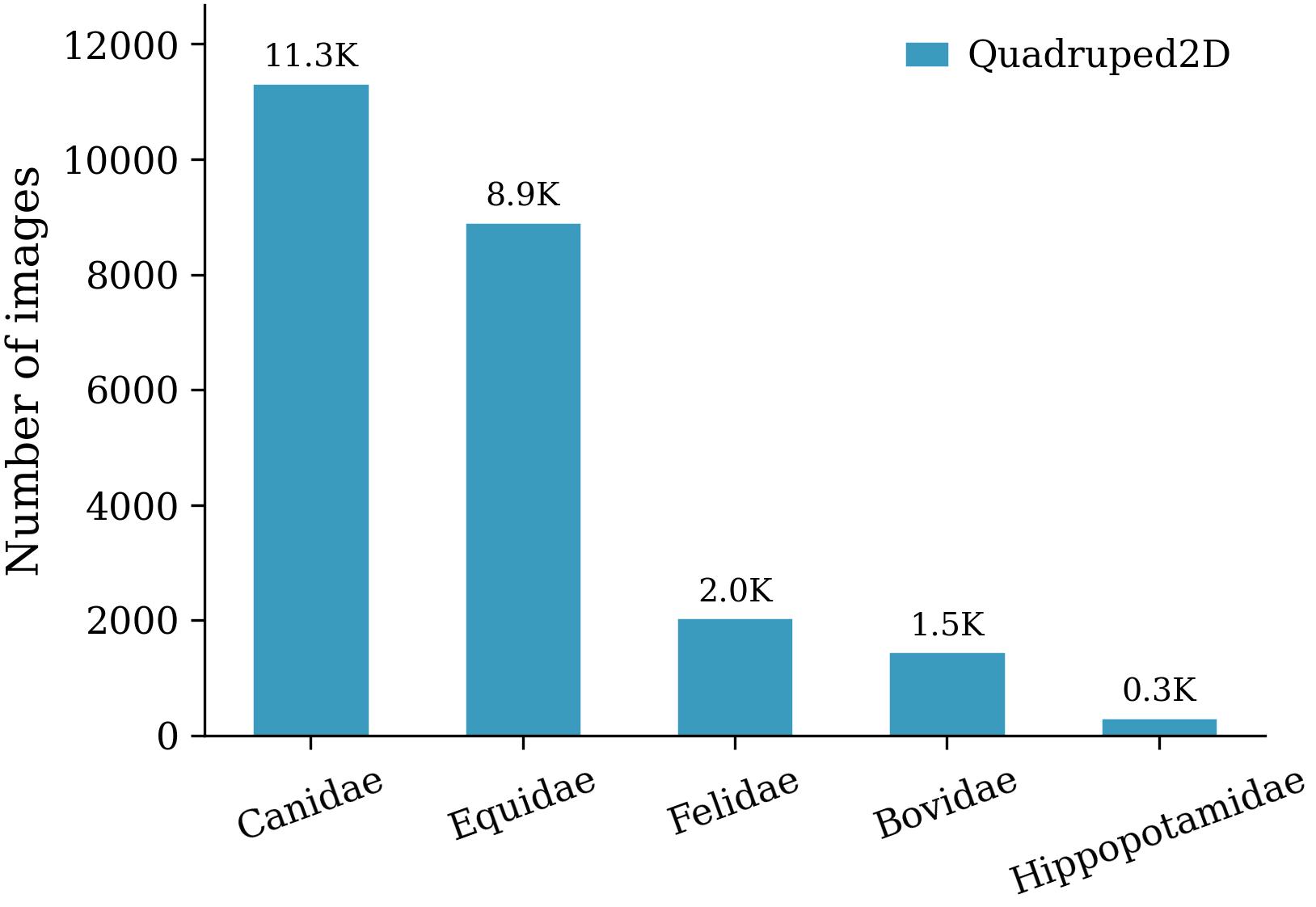}
        \caption{Quadruped2D.}
    \end{subfigure}
    \caption{\textbf{Family-level image distributions of the datasets used in this work.}}
    \label{fig:categornimal3d}
\end{figure}

To characterize the imbalance of animal categories in existing datasets, we analyze the taxonomic distribution of the existing datasets.
First, we examine the distribution of samples at the family level. Fig.~\ref{fig:categornimal3d} shows the number of instances for five animal families in three widely used datasets, Animal3D \cite{xu2023animal3d}, CtrlAni3D \cite{lyu2025animer}, and Quadruped2D \cite{ye2024superanimal}.
As illustrated in the figure, the distribution across families is highly uneven, with Equidae and Canidae dominating the dataset while others contain substantially fewer samples. This imbalance suggests that current datasets may bias learning algorithms toward more frequently observed animal families.

\begin{figure}[t]
    \centering
    \includegraphics[width=1.0\linewidth]{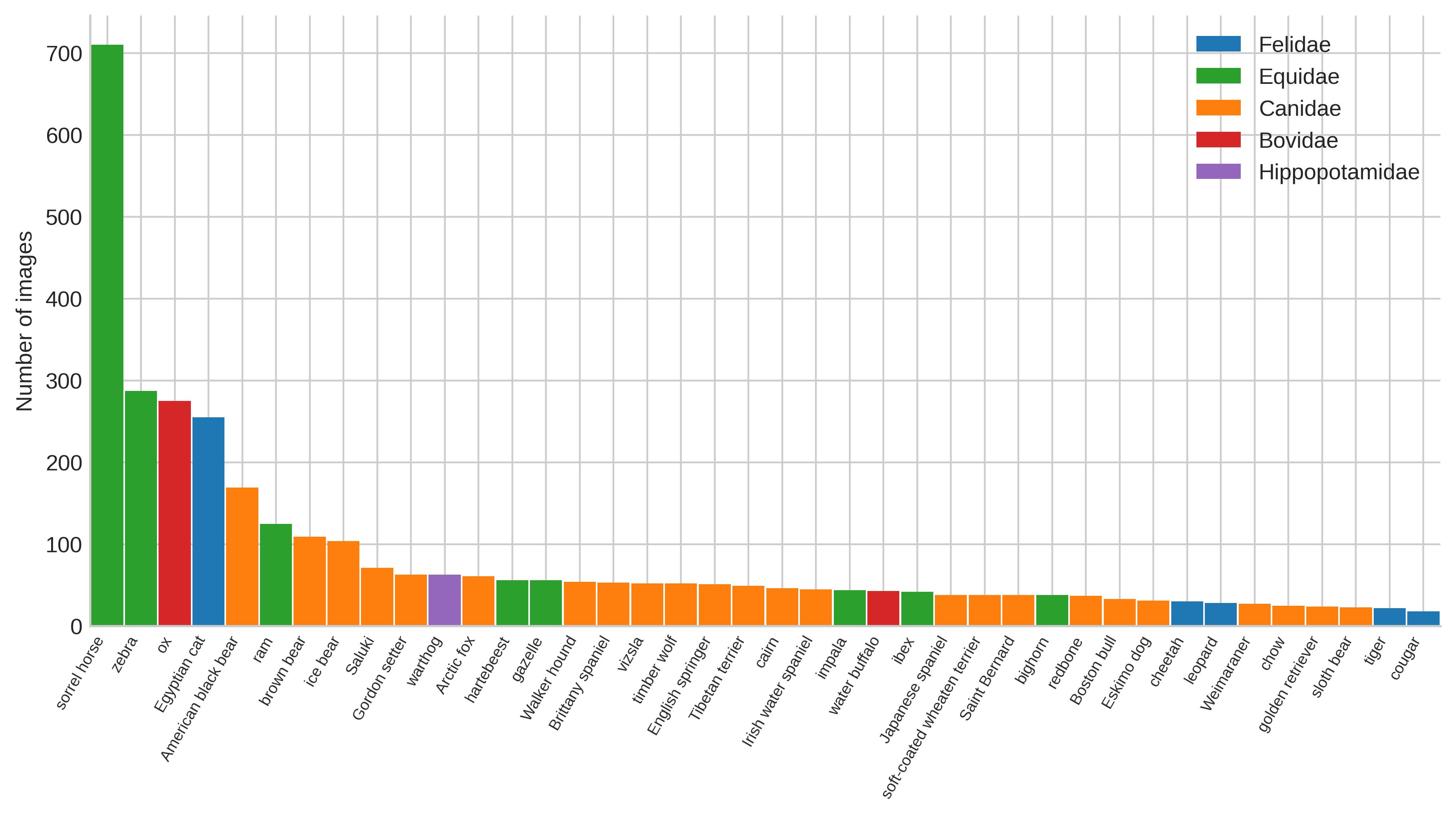}
    \caption{\textbf{Species-level image distribution of the Animal3D dataset across 40 categories, sorted by total count in descending order.} Bar colors indicate the taxonomic family: \textcolor[HTML]{1f77b4}{Felidae}, \textcolor[HTML]{2ca02c}{Equidae}, \textcolor[HTML]{ff7f0e}{Canidae}, \textcolor[HTML]{d62728}{Bovidae}, and \textcolor[HTML]{9467bd}{Hippopotamidae}. The dataset contains 3{,}385 images in total and exhibits significant class imbalance, with the most populated category (\textit{sorrel horse}, 710 images) containing $\sim$39$\times$ more images than the least populated (\textit{cougar}, 18 images).}
    \label{fig:category_distri_animal3d}
\end{figure}

We further analyze the distribution at the species level using the Animal3D \cite{xu2023animal3d} dataset. As shown in Fig.~\ref{fig:category_distri_animal3d}, the dataset contains 40 animal species, but the number of samples per species varies significantly.
The resulting distribution exhibits a long-tailed pattern, where a small subset of species accounts for the majority of samples, while many species have relatively few instances. Such an imbalance may hinder the learning of robust and generalizable animal representations.

\section{Additional Implementation Details} 
\subsection{Details of network training}
We implement PRIMA in PyTorch and extract the BioCLIP \cite{stevens2024bioclip} embeddings with the \texttt{vit\_b16\_pn\_bioclip-v1} model. The bio-encoder takes the resulting 512-dimensional BioCLIP feature representation and projects it into a 1280-dimensional bio-token space, while the ViT image encoder produces a sequence of visual tokens of size \(192 \times 1280\). Conditioned on these two token sequences, the keypoint-aware decoder progressively refines the SMAL parameters through an Iterative Error Feedback (IEF) loop, where each IEF iteration passes through \(L = 6\) decoder layers, and both the 2D and 3D keypoint estimates are refined at every layer.

The network is trained in two stages. In the warm-up stage, it is pre-trained exclusively on the Animal3D dataset for 500 epochs, so that the model first learns plausible shapes and poses; the full set of training data is then introduced in the second stage. The number of IEF iterations follows the same schedule: we use \(N = 3\) iterations during the warm-up stage and reduce it to \(N = 1\) afterwards, which improves computational efficiency once the predictions are already close to plausible configurations. We further adopt different learning rates for the encoder and the remaining network components, optimizing the encoder with a smaller learning rate of \(3.75\times10^{-7}\) and the remaining parameters with a learning rate of \(3.75\times10^{-6}\). All parameters are updated with AdamW \cite{loshchilov2017adamw} using a batch size of 48 and a linear learning-rate decay schedule. Training stops after 1200 epochs in total, at which point the training loss is observed to stabilize, amounting to an overall training time of 46 hours on a single NVIDIA A100 GPU.

\subsection{Details of the adaptation process}

Our proposed adaptation approach optimizes the SMAL parametric model's shape and pose parameters using only ground-truth 2D keypoint annotations.
During the adaptation process, we optimize the parameters within the SMAL regression head, including learnable query tokens and regression MLPs for shape, pose, and camera parameters. The optimization is performed using the Adam optimizer with a learning rate of $1 \times 10^{-6}$ 
for 10 to 60 iterations per sample, depending on the baseline performance.
The optimization objective focuses on SMAL-reconstructed and auxiliary 2D keypoints, ensuring gradients effectively reach shape and pose parameters. The loss weights are set to:
\begin{equation}
\lambda_{\text{aux}} = 5.0, \quad \lambda_{\text{reproj}} = 0.5 .
\end{equation}

In terms of runtime, baseline inference takes approximately 0.05 s per image on an NVIDIA A100 GPU. With the adaptation process enabled, the runtime increases to approximately 0.35 s per image for 10 optimization iterations.

\section{Additional Results}

\subsection{\textcolor{blue}{Comparisons with non-SMAL-based methods}}
\textcolor{blue}{
We further compare PRIMA with recent animal reconstruction methods that do not rely on SMAL as their underlying parametric shape model. These methods provide complementary comparisons to the SMAL-based approaches evaluated in the main paper, although their output representations and evaluation protocols differ from ours.}

\begin{figure}
    \centering
    \includegraphics[width=0.92\linewidth]{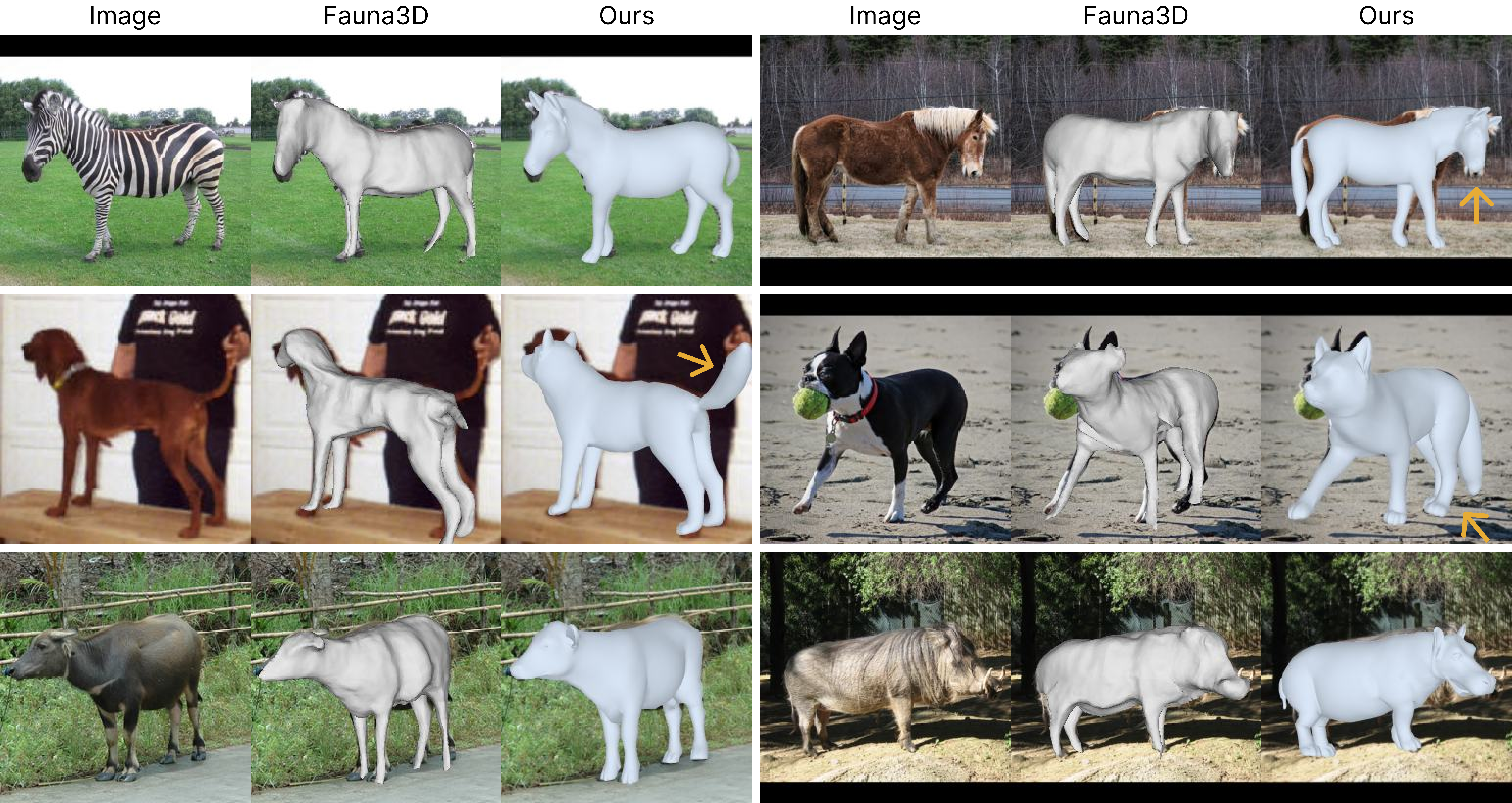}
    \caption{\textbf{\textcolor{blue}{Qualitative comparisons with Fauna3D.}} \textcolor{blue}{Orange arrows indicate better pose recovery in PRIMA.}}
    \label{fig:fauna}
\end{figure}

\textcolor{blue}{\textbf{Comparison with Fauna3D.}
We compare PRIMA with Fauna3D \cite{li20243dfauna} on Animal3D using bbox-normalized PCK@0.1, following the evaluation protocol of Fauna3D. As shown in Tab.~\ref{tab:fauna}, Fauna3D achieves 90.1, while our model achieves 91.5.
Since PCK primarily measures 2D alignment and does not fully reflect mesh quality, we additionally provide qualitative comparisons in Fig.~\ref{fig:fauna}. At the representation level, its model-free meshes tend to exhibit coarser local geometry, whereas the SMAL representation used by PRIMA provides a consistent topology with more clearly defined anatomical structures. Beyond this representation difference, PRIMA more faithfully captures instance-specific body proportions and articulated configurations across diverse animal morphologies, consistent with its higher PCK.}

\begin{figure}
    \centering
    \includegraphics[width=0.92\linewidth]{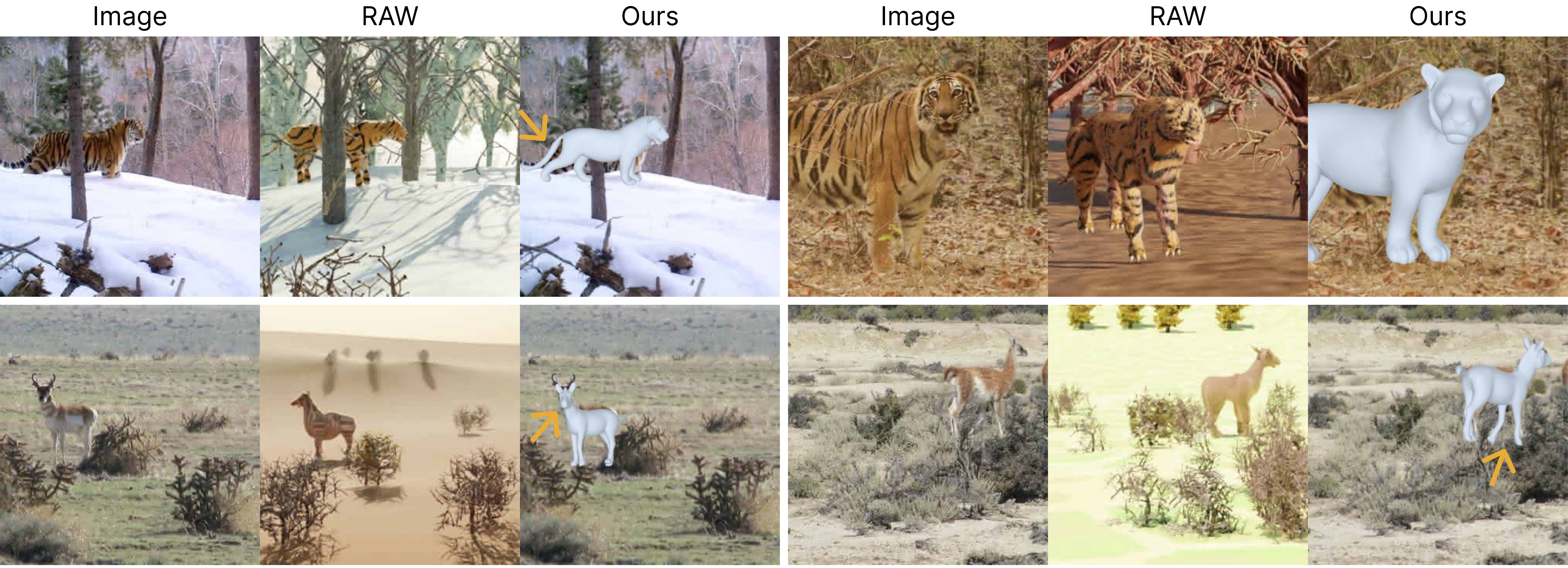}
    \caption{\textbf{\textcolor{blue}{Qualitative comparisons with RAW. }}\textcolor{blue}{Orange arrows indicate better pose recovery in PRIMA.}} 
    
    \label{fig:raw}
\end{figure}

\textcolor{blue}{\textbf{Comparison with RAW.}
RAW \cite{kulits2025raw} targets compositional animal--environment reconstruction, rather than instance-level articulated animal mesh recovery. Its representation and evaluation protocol are therefore incompatible with our metrics, precluding a direct quantitative comparison. We instead provide qualitative comparisons in Fig.~\ref{fig:raw}, using input images from the RAW demo. 
While RAW emphasizes holistic scene reconstruction, PRIMA focuses on instance-specific articulated reconstruction and more closely follows the input pose in these examples, e.g., the tiger's leg configuration and the antelope's head orientation.}

\begin{table}[t]
\centering
\small
\setlength{\tabcolsep}{4pt}
\renewcommand{\arraystretch}{1.1}
\begin{tabular}{ccc}
\hline
Method   &Fauna3D & Ours\\ \hline
PCK@0.1 $\uparrow$   &  90.1      & \textbf{91.5} 
\\ \hline
\end{tabular}
\caption{\textbf{\textcolor{blue}{Comparison with Fauna3D on Animal3D.}} We use bbox-normalized PCK@0.1 here (as in Fauna3D), while the main paper reports mask-normalized PCK.}
\label{tab:fauna}
\end{table}

\begin{figure*}
    \centering
    \includegraphics[width=0.84\linewidth]{fig/supp_qualitative_more.pdf}
    \caption{\textbf{Qualitative comparisons on the Animal3D and Quadruped2D dataset.}\textcolor{blue}{ Orange outlines highlight representative cross-family shape outliers, complementing the family-level shape-space analysis quantified by ARI.}}
    \label{fig:supp_quali}
\end{figure*}

\subsection{Qualitative comparisons with state-of-the-art methods}

We present additional qualitative comparisons with SOTA methods on the Animal3D and the Quadruped2D dataset. As shown in Fig.~\ref{fig:supp_quali}, the first four rows depict instances from the Bovidae family. In these examples, other SOTA methods hallucinate shapes resembling canines, whereas our approach more faithfully reconstructs bovine morphology. 
Row 5 shows a sheep example: AniMer captures overall body shape but has less accurate pose, while AniMer+ incorrectly predicts a feline-like geometry. 
In row 6, only our method reconstructs a shape consistent with the appearance of a bear. Similarly, in the Quadruped2D dataset, our results show a closer alignment between shape and animal species.

\subsection{Family-wise quantitative comparisons}

To assess method performance across varying morphologies, we perform family-wise comparisons on the Animal3D dataset.
Tab.~\ref{tab:quan_categ} presents family-wise quantitative comparisons on the Animal3D dataset across five animal families. Overall, our method demonstrates strong and consistent performance across categories, achieving the best PA-MPVPE results in four families (Equidae, Canidae, Felidae, and Bovidae). Notably, the improvement on Bovidae is particularly significant, suggesting that our method better handles animals with larger body variations.
For Hippopotamidae, AniMer achieves the lowest errors. We attribute this to the limited number of samples (6 samples) and the distinct body proportions of this category compared with other animals in the dataset. Nevertheless, our method still delivers competitive performance, demonstrating cross-category generalization. 


\begin{table}[h]
\centering
\small
\renewcommand{\arraystretch}{1.1}
\begin{tabular}{c|cc|cc|cc|cc|cc}
\toprule
 & \multicolumn{2}{c|}{Equidae} 
 & \multicolumn{2}{c|}{Canidae} 
 & \multicolumn{2}{c|}{Felidae} 
 & \multicolumn{2}{c|}{Bovidae} 
 & \multicolumn{2}{c}{Hippopotamidae} 
\\
Method 
 & PAJ$\downarrow$ & PAV$\downarrow$  
 & PAJ$\downarrow$ & PAV$\downarrow$ 
 & PAJ$\downarrow$ & PAV$\downarrow$ 
 & PAJ$\downarrow$ & PAV$\downarrow$ 
 & PAJ$\downarrow$ & PAV$\downarrow$ \\
\midrule
AniMer \cite{lyu2025animer}   & 72.2 & 77.8 & 69.5 & 72.1 & 118.6 & 122.1 & 78.2 & 78.2 & \textbf{84.2} & \textbf{103.9}   \\
AniMer+ \cite{an2025animer+}   & 75.1 & 81.1 & \textbf{68.5} & 72.9 & 118.0 & 120.1 & 80.0 & 80.3 & 98.4 & 118.5\\
\midrule
Ours   & \textbf{70.3} & \textbf{76.0} & \underline{68.9} & \textbf{71.7} & \textbf{115.3} & \textbf{118.2} & \textbf{72.2} & \textbf{73.2} & \underline{91.3} & \underline{110.4}  \\
\bottomrule
\end{tabular}
\caption{\textbf{Family-wise quantitative comparison on the Animal3D dataset.} \textbf{Bold}: best. \underline{Underline}: second best.}
\label{tab:quan_categ}
\end{table}

\subsection{More qualitative results on our Quadruped3D}
We provide additional qualitative results on our Quadruped3D dataset, encompassing a diverse set of animal species, viewpoints, and articulated poses, as illustrated in Fig.~\ref{fig:supp_qua3d}. The refined pGT, depicted in green, improves spatial alignment, even under challenging conditions such as extreme poses (e.g., a dog lying on its belly, a cat climbing a tree) and occlusions.

\begin{figure}[h]
    \centering
    \includegraphics[width=0.95\linewidth]{fig/supp_quadruped3d.pdf}
    \caption{\textbf{More examples on Quadruped3D pGT. }}
    \label{fig:supp_qua3d}
\end{figure}

\subsection{More qualitative results on Animal Kingdom dataset}
\begin{figure}
    \centering
    \includegraphics[width=0.98\linewidth]{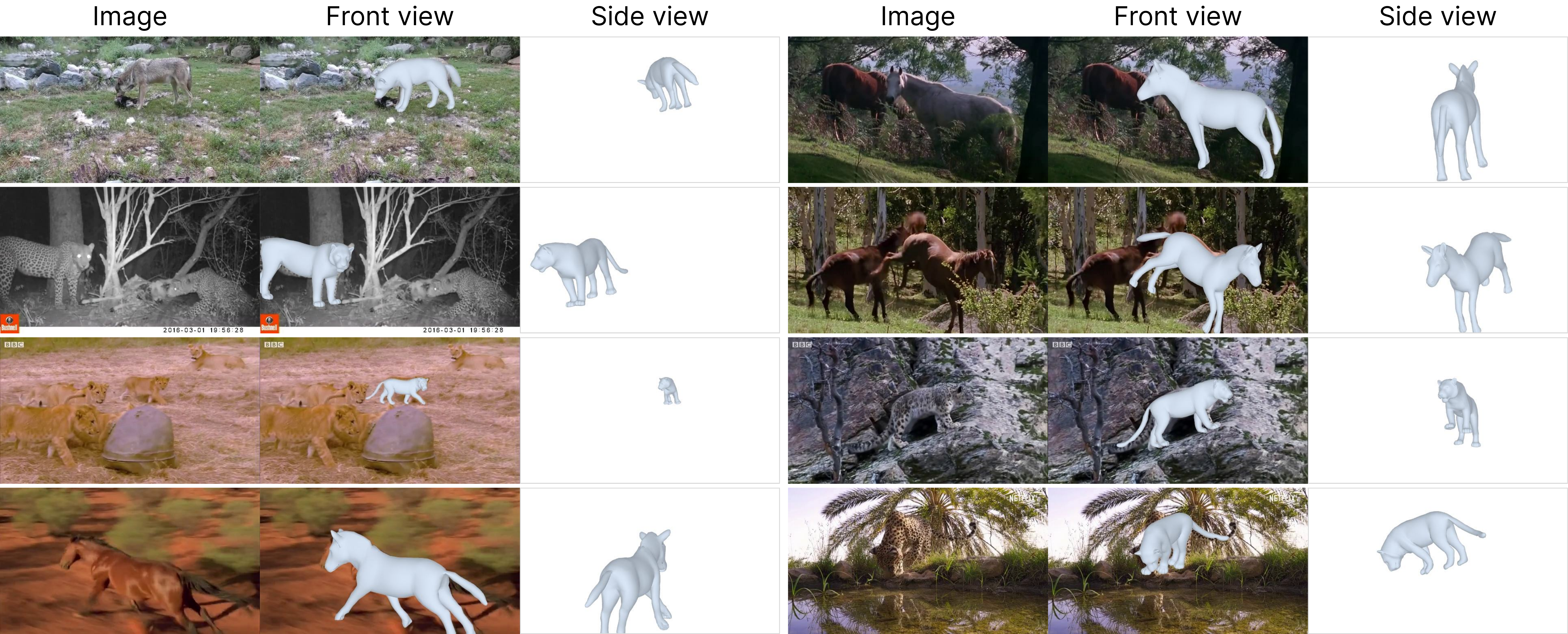}
    \caption{\textbf{\textcolor{blue}{Qualitative results on the Animal Kingdom dataset.} }}
    \label{fig:supp_kingdom}
\end{figure}

We present qualitative results on the Animal Kingdom dataset, which consists of challenging in-the-wild imagery. As illustrated in Fig.~\ref{fig:supp_kingdom}, PRIMA exhibits consistently robust performance even under challenging imaging conditions, including occlusions (second example in the first row), motion blur (first example in the fourth row), and artifacts induced by camera traps (first example in the second row).


\section{\textcolor{blue}{Analysis and validation of the adaptation process}} 
\begin{figure}[h]
    \centering
    \includegraphics[width=0.65\linewidth]{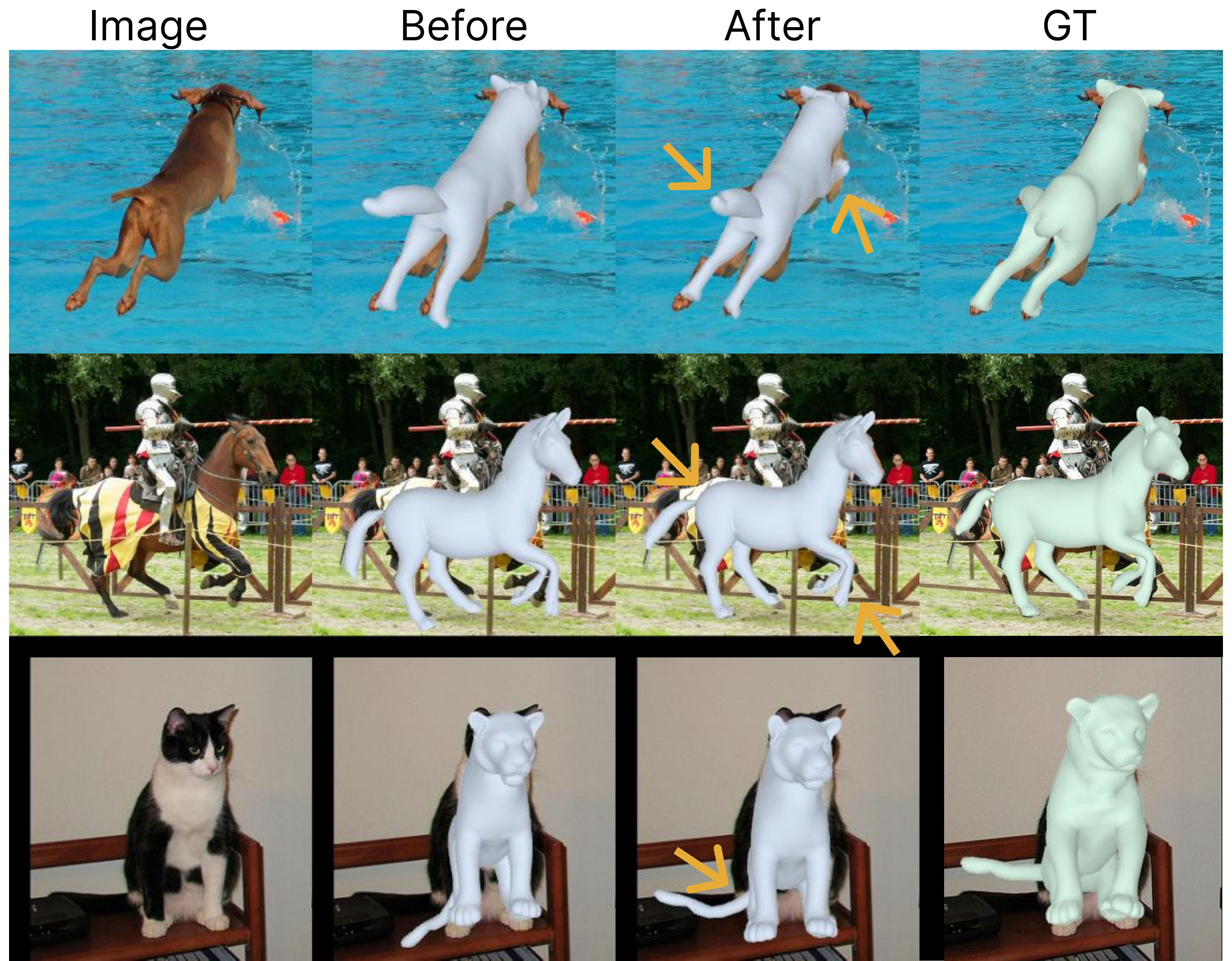}
    \caption{\textcolor{blue}{\textbf{Qualitative analysis of the adaptation process on Animal3D.}}
Orange arrows highlight improved alignment with the input image after adaptation.}
    \label{fig:tta}
\end{figure}

\textcolor{blue}{To directly assess whether our adaptation procedure can generate geometrically valid pseudo-3D labels from 2D supervision, we conduct a controlled experiment on Animal3D. During adaptation, we use only the ground-truth 2D keypoints and do not expose the model to the corresponding 3D ground truth. We treat the adapted meshes as pseudo-GT, exactly following the procedure used to construct Quadruped3D, and evaluate them against the held-out real 3D annotations.}
We focus on high-error cases from the Animal3D dataset, as shown in Fig.~\ref{fig:tta}. These cases typically involve out-of-domain poses and significant self-occlusion, highlighting the challenges addressed by our method. Quantitative comparisons are reported in Tab.~\ref{tab:quan_tta}. 
All cases show improvements in both 2D and 3D. 

\begin{table}[h]
\centering
\small
\renewcommand{\arraystretch}{1.1}
\begin{tabular}{c|cc|cc|cc}
\toprule
 & \multicolumn{2}{c|}{PAJ$\downarrow$} 
 & \multicolumn{2}{c|}{PAV$\downarrow$} 
 & \multicolumn{2}{c}{AUC$\uparrow$}    \\
Instance & Before & After & Before & After & Before & After \\
\midrule
Dog jumping    &  133.1 & \textbf{117.6} & 114.2 & \textbf{80.6} & 81.9 & \textbf{90.1}  \\
Horse running     & 52.1 & \textbf{46.0} & 53.2 & \textbf{52.9} & 90.5 & \textbf{93.7}  \\
Cat sitting     & 113.6 & \textbf{88.3} & 77.2  & \textbf{67.2} & 86.5 & \textbf{94.6} \\
Avg over Animal3D &75.3 & \textbf{71.7}   &79.5 & \textbf{75.9}   & 89.6 & \textbf{91.5} \\
\bottomrule
\end{tabular}
\caption{
\textbf{\textcolor{blue}{Quantitative validation of the adaptation process on Animal3D.}}
\textcolor{blue}{
We compare performance before and after adaptation using 2D and 3D metrics on three representative
examples and the overall Animal3D test set.
}
\textbf{Bold} indicates improvement after adaptation.
\textcolor{blue}{
The model is trained without Quadruped3D; Animal3D 3D annotations are used
only for evaluation and are not used during adaptation.
}}

\label{tab:quan_tta}
\end{table}

We analyze the influence of optimization iterations on the adaptation performance using the horse-running scenario as a case study. We report results in terms of the relative improvement with respect to the baseline obtained without adaptation (i.e., 0 optimization steps). For error metrics (PA-MPJPE and PA-MPVPE), the improvement corresponds to the relative error reduction compared to the baseline, while for AUC it represents the relative increase over the baseline.

\begin{figure}[h]
    \centering
    \includegraphics[width=\linewidth]{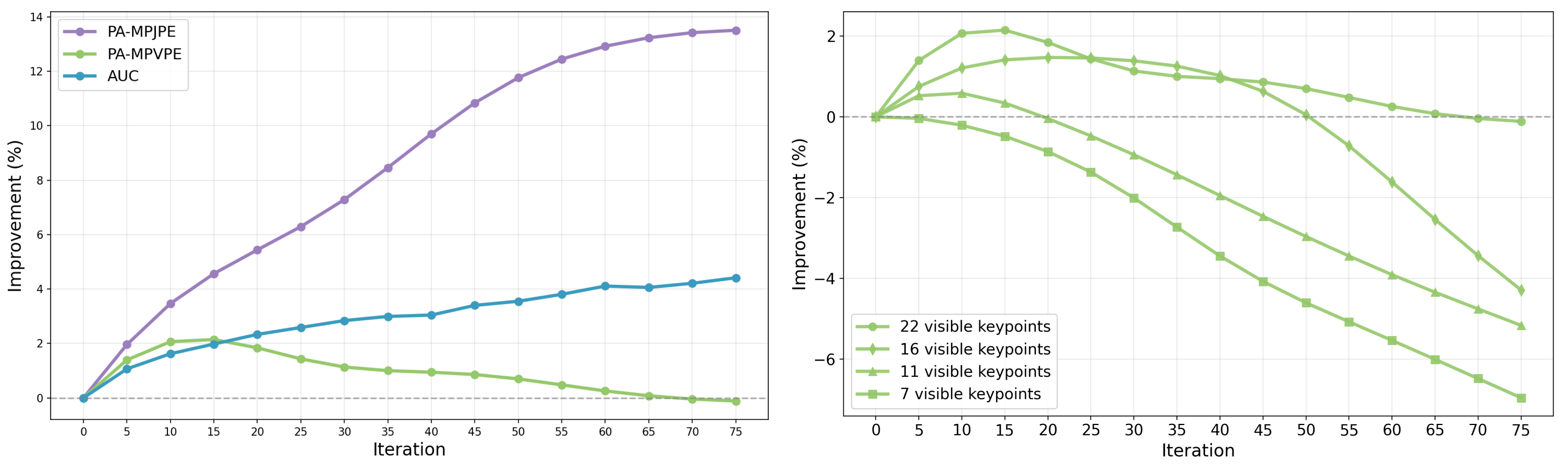}
    \caption{\textbf{Effect of optimization iterations and visible keypoints in the horse-running case.} 
    Left: Relative improvement of PA-MPJPE, PA-MPVPE, and AUC as the number of optimization iterations increases. 
    Right: Relative PA-MPVPE improvement under different numbers of visible keypoint conditions as the optimization iteration increases.}
    \label{fig:supp_tta_iters}
\end{figure}

As shown in the left panel of Fig.~\ref{fig:supp_tta_iters}, increasing the number of optimization iterations consistently improves PA-MPJPE and AUC, indicating that the adaptation helps refine the pose estimation by better aligning the 3D prediction with the 2D observations. In contrast, the PA-MPVPE improvement initially increases, peaking at approximately 15 steps, after which it gradually declines. When the number of optimization iterations grows large (around 70 steps), performance degrades and even drops below the no-adaptation baseline. This suggests that excessive optimization overfits the 2D observations and degrades mesh reconstruction, especially without using silhouette information, unlike the SMALify \cite{biggs2018creatures} optimization process.

We also investigate how the number of 2D keypoints influences the adaptation process. In the horse-running scenario, the original annotation provides 22 visible keypoints. To evaluate the robustness to keypoint sparsity, we randomly mask $30\%$, $50\%$, and $70\%$ of the keypoints, resulting in 16, 11, and 7 visible keypoints, respectively. The right panel of Fig.~\ref{fig:supp_tta_iters} shows the PA-MPVPE improvement across different visible keypoint conditions. The number of available keypoints significantly affects the stability of the adaptation process. When sufficient keypoints are available, a small number of optimization steps leads to moderate performance gains, while excessive iterations gradually degrade the results. In contrast, when only 7 keypoints are visible, the performance quickly deteriorates as the number of iterations increases. These results show that dense, reliable keypoint observations are crucial for stable, effective mesh adaptation. 

\section{Limitation and future work}
\textcolor{blue}{Although the BioCLIP-derived taxonomic prior is effective for family-level generalization, it mainly encodes species-level taxonomic semantics} and may miss fine-grained anatomical details (e.g., local body proportions, facial structures, breed-specific traits). As a result, reconstruction quality can decline for subtle within-species geometric variations. 

\begin{figure}[h]
    \centering
    \includegraphics[width=0.9\linewidth]{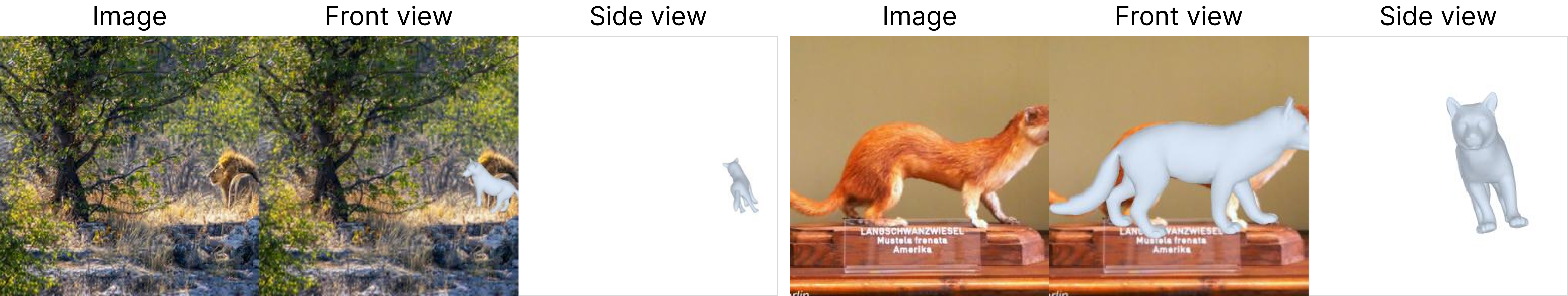}
    \caption{\textbf{\textcolor{blue}{Failure cases.}}}
    \label{fig:failure}
\end{figure}

\textcolor{blue}{Fig.~\ref{fig:failure} further illustrates two representative failure cases. For distant animals occupying few pixels, limited visual evidence and pose ambiguity can affect both taxonomic cues and pose estimation, as illustrated by the lion being reconstructed with a dog-like shape and inaccurate pose.
Moreover, animals whose morphology falls outside the coverage of SMAL and our training data remain challenging. For example, PRIMA recovers a plausible pose for the unseen weasel but cannot faithfully represent its distinctive elongated body shape.}

\textcolor{blue}{Future work could incorporate dense correspondence representations, such as DensePoseCSE \cite{Neverova2020ContinuousSurfaceEmbeddings,Neverova2021UniversalCanonicalMaps}, for finer anatomical guidance, and explore more expressive parametric or mesh-based shape representations (e.g., SMAL+) to better accommodate diverse and unseen animal morphologies.}


\twocolumn

{
    \small
    \bibliographystyle{ieeenat_fullname}
    \bibliography{main}
}
\end{document}